\documentclass[acmtog]{acmart}

\usepackage{booktabs} 
\usepackage{comment}
\usepackage{bm}
\usepackage{multirow}
\usepackage{placeins}
\usepackage{svg}
\usepackage{mdframed}
\usepackage{xcolor} 
\usepackage{array}  
\usepackage[ruled]{algorithm2e} 
\usepackage{subfig}
\usepackage{footnote}

\SetAlFnt{\small}
\SetAlCapFnt{\small}
\SetAlCapNameFnt{\small}
\SetAlCapHSkip{0pt}
\usepackage{enumitem}
\usepackage{xcolor}
\acmJournal{TOG}

\newcommand{\secref}[1]{\S~\ref{sec:#1}}
\newcommand{\tabref}[1]{Tab.~\ref{tab:#1}}
\newcommand{\figref}[1]{Fig.~\ref{fig:#1}}
\renewcommand{\eqref}[1]{Eq.~(\ref{eq:#1})}

\newcommand{\eg}{\emph{e.g., }}

\newcommand{\mbA}{\ensuremath{\mathbf{A}}}

\newcommand{\mbx}{\ensuremath{\mathbf{x}}}
\newcommand{\mbn}{\ensuremath{\mathbf{n}}}

\newcommand{\mby}{\ensuremath{\mathbf{y}}}
\newcommand{\mbz}{\ensuremath{\mathbf{z}}}
\newcommand{\mbv}{\ensuremath{\mathbf{v}}}

\begin{document}
	\title{Online Neural Path Guiding with Normalized Anisotropic Spherical Gaussians}
	
	\author{Jiawei Huang}
	\affiliation{%
		\institution{Chuzhou University}
		\city{Chuzhou}
		\country{China}}
        \affiliation{%
		\institution{Void Dimensions}
		\country{China}}
	\author{Akito Iizuka}
	\affiliation{%
		\institution{Tohoku University}
		\city{Sendai}
		\country{Japan}
	}
	\author{Hajime Tanaka}
	\affiliation{%
		\institution{Tohoku University}
		\city{Sendai}
		\country{Japan}}
	\author{Taku Komura}
	\affiliation{
		\institution{The University of Hong Kong}
		\country{Hong Kong}
	}
        \affiliation{
		\institution{Tohoku University}
		\city{Sendai}
		\country{Japan}
	}
	\author{Yoshifumi Kitamura}
	\affiliation{%
		\institution{Tohoku University}
		\city{Sendai}
		\country{Japan}
	}

	\begin{abstract}
		Importance sampling techniques significantly reduce variance in physically-based rendering. In this paper we propose a novel online framework to learn the spatial-varying distribution of the full product of the rendering equation, with a single small neural network using stochastic ray samples. The learned distributions can be used to efficiently sample the full product of incident light.
		To accomplish this, we introduce a novel closed-form density model, called the Normalized Anisotropic Spherical Gaussian mixture, that can model a complex light field with a small number of parameters and that can be directly sampled.
		Our framework progressively renders and learns the distribution, without requiring any warm-up phases.
		With the compact and expressive representation of our density model, 
		our framework can be implemented entirely on the GPU, allowing it to produce high-quality images with limited computational resources.
        The results show that our framework outperforms existing neural path guiding approaches and achieves comparable or even better performance than state-of-the-art online statistical path guiding techniques.
		
	\end{abstract}
	
	\begin{teaserfigure}
  \begin{tabular}{rccccc}
  			&NASG (Ours)&NIS&PPG&PAVMM&Reference
			\\
  			&\textbf{0.042}&0.043&0.052&0.048&MAPE
			\\
			\begin{minipage}{0.45\linewidth}
                \centering
				\includegraphics[trim = 150 0 150 0, clip,height=5.7cm]{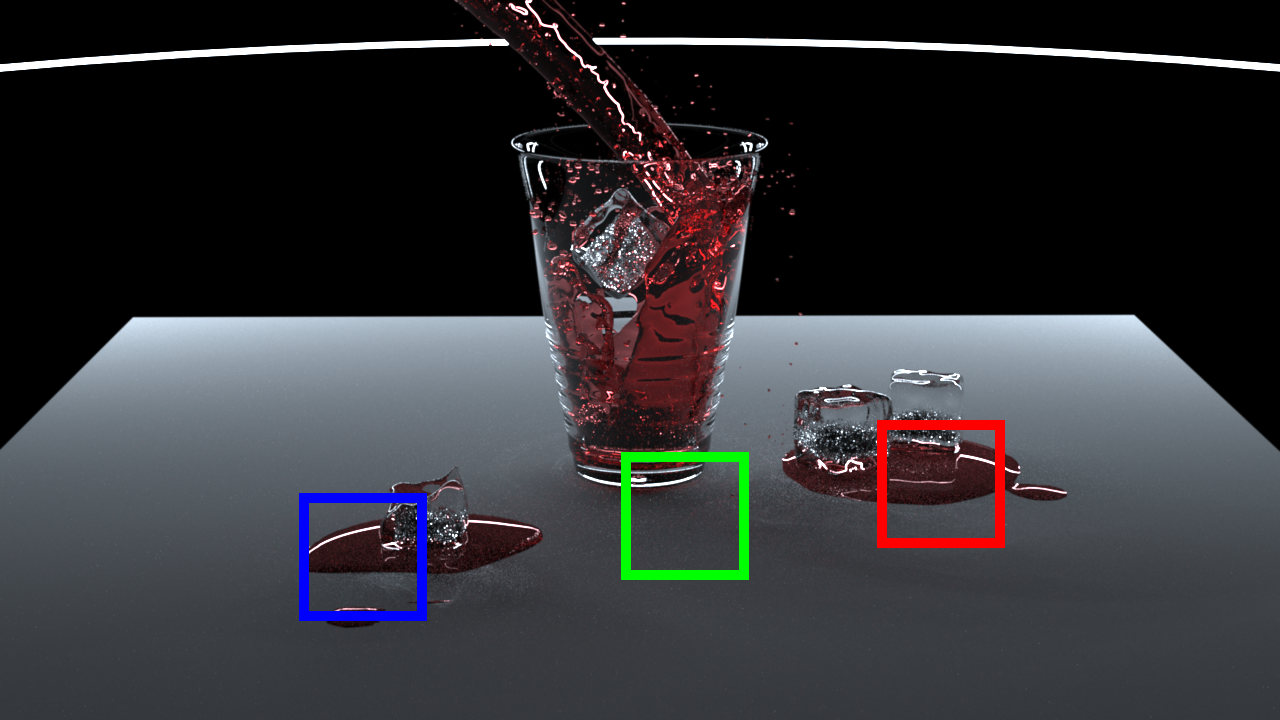}
			\end{minipage}
                &
			\begin{minipage}{0.08\linewidth}
				\includegraphics[height=1.9cm]{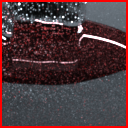}
				\newline
				\includegraphics[height=1.9cm]{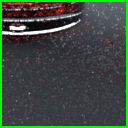}
				\newline
				\includegraphics[height=1.9cm]{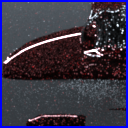}
			\end{minipage}
                &
			\begin{minipage}{0.08\linewidth}
				\includegraphics[height=1.9cm]{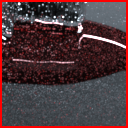}
				\newline
				\includegraphics[height=1.9cm]{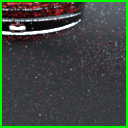}
				\newline
				\includegraphics[height=1.9cm]{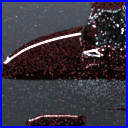}
			\end{minipage}
                &
			\begin{minipage}{0.08\linewidth}
				\includegraphics[height=1.9cm]{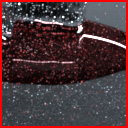}
				\newline
				\includegraphics[height=1.9cm]{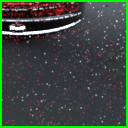}
				\newline
				\includegraphics[height=1.9cm]{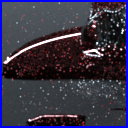}
			\end{minipage}
                &
			\begin{minipage}{0.08\linewidth}
				\includegraphics[height=1.9cm]{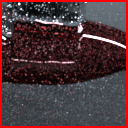}
				\newline
				\includegraphics[height=1.9cm]{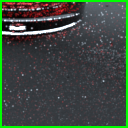}
				\newline
				\includegraphics[height=1.9cm]{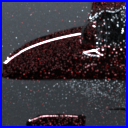}
			\end{minipage}
                &
			\begin{minipage}{0.08\linewidth}
				\includegraphics[height=1.9cm]{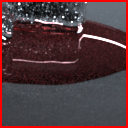}
				\newline
				\includegraphics[height=1.9cm]{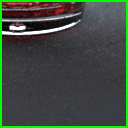}
				\newline
				\includegraphics[height=1.9cm]{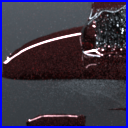}
			\end{minipage}
                \\
  \end{tabular}
		\caption{\label{fig:teaser}
			\textsc{Ajar} scene rendered using proposed online path guiding framework and previous path guiding techniques, including Neural Importance Sampling (NIS) \cite{MUL19}, Practical Path Guiding (PPG) \cite{PPG19}, and the Robust Fitting of Parallax-Aware Mixtures for Path Guiding (PAVMM) \cite{PAPG}, with the same 512 sample budget. Our framework uses a single multilayer perceptron (MLP) to learn the continuous distribution of the full scattered radiance product, represented as normalized anisotropic spherical Gaussian mixtures, to achieve lower sampling variance.  }
	\end{teaserfigure}
	
	\maketitle
	
	%
	%

	
	%
	%

	\section{Introduction}
	Unbiased physically-based rendering (PBR) is achieved by launching a light transport simulation to solve the rendering equation with Monte Carlo methods. Over the past few decades, unidirectional path tracing has become the dominant method of PBR in the film and design industries due to its flexibility and simplicity. To reduce the variance of the Monte Carlo integral efficiently, many importance sampling methods have been proposed (\eg \cite{VEACH98, SA13, hart2020practical}). However, currently, only some components in the rendering equation (typically, the BSDF term, or direct lighting) can be importance-sampled. When indirect lighting is dominant, the sampling efficiency becomes poor.
	
	Path guiding is a promising genre of importance sampling approaches to overcome this challenge~\cite{VORBA14,MUL17}. A path guiding method usually learns distributions over the 3D scene that fit the rendering equation more closely, either during the rendering process (online) or with precomputation (offline). 
	Then, the path tracer can sample the scattering directions using the learned distribution to reduce the variance. 
	Previous approaches~\cite{VORBA14,MUL17}
	have partitioned the 3D space and approximated the incident radiance of each zone using 
	statistical methods (typically, histogramming or expectation-maximization) 
	instead of modeling the whole product of the rendering equation for each shading point; consequently, they could only learn one distribution for each spatial partition and usually suffered from the parallax issue.

	Recently, neural networks are starting to be used for learning spatially varying, per-shading-point product distribution of the rendering equation \cite{MUL20,SHANDONG21,ZHU21}. 
	Ideally, an explicit model that directly fits 
	the continuous product distribution over 3D space is desired; however, this has been difficult due to the lack of (1) a representation that can reconstruct both high- and low-frequency features with a small number of parameters and (2) a closed-form density model that can provide an analytical solution for integrating the distribution for normalization.
	As a result, existing models 
	either (1) learn an implicit model that only provides a mapping between a uniform distribution and the shading-point product distribution, but with expensive computation ~\cite{MUL20};  
	(2) learn a coarse, low-resolution distribution with limited accuracy~\cite{SHANDONG21}; or (3) require costly offline precomputation~\cite{ZHU21}. 
 
	In this paper, we propose a novel, neural-network-based path guiding framework that learns an explicit spatial-varying distribution of scattered radiance over the surfaces in a 3D scene online. 
    Using the shading point's 3D position and auxiliary information as input, our network estimates the distribution of full scattered radiance product, which can be used for efficient importance sampling of light transport.
    Our key insight is a novel closed-form density representation, namely the Normalized Anisotropic Spherical Gaussian mixture (NASG). The NASG is a $5$-parameter anisotropic distribution model that comes with an easy-to-compute integral and a closed-form sampling algorithm, making it more feasible for Monte Carlo rendering compared to Kent distribution. We also show that normalization is an important factor for successful learning of scattered radiance distribution with sparse online training samples. The simplicity and expressiveness of NASG, along with its analytical normalization formula, lead to successful online learning of the complex radiance distribution via a tiny multilayer perceptron (MLP).

	Our framework is robust enough to handle a variety of lighting setups, ranging from normal indirectly illuminated scenes to caustics. We also propose the corresponding training workflow, including specialized loss function and online acquisition of training samples. 
Because our network can be easily stored and executed in parallel on a GPU, we are able to integrate it with a wavefront path tracer. In this way, we can implement a high-performance, neural-guided, unidirectional path tracer on a single GPU, making neural guiding affordable for conventional personal computers.
    Our framework outperforms existing neural path guiding approaches both in terms of sampling efficiency and raw performance, and it provides comparable or even better performance than state-of-the-art   
    statistical methods.

	The contributions of this paper can be summarized as
	\begin{itemize}
		\item Normalized Anisotropic Spherical Gaussian mixture, a novel density model that can effectively represent spatial-varying distribution for path tracing, and
		\item A novel, lightweight, online neural path guiding framework that can be integrated with either CPU- or GPU-based conventional production path tracers.
	\end{itemize}

	\section{Related Work}
	\subsection{Physically-based Rendering and Path Guiding}
	The radiance of a shading point can be calculated using the rendering equation \cite{KAJIYA86}:
	\begin{equation}
    \label{eq:rendering}
	L(\mbx,\omega_o) = L_e(\mbx) + \int_{S^2}f_s(\mbx,\omega_i, \omega_o)L_i(\mbx, \omega_i) |\cos\theta_i| d\omega_i,
	\end{equation}
which consists of the emitted radiance \(L_e(\mbx)\) at the point and the integral over the sphere \(S^2\), which aggregates the contributions of the incident radiance \(L_i(\mbx, \omega_i)\) from all directions \(\omega_i\). For each \(\omega_i\), the product of incident radiance, the bidirectional scattering distribution function (BSDF) \(f_s(\mbx,\omega_i, \omega_o)\), and the geometry term \(|\cos\theta_i|\) signify its contribution to the outgoing radiance.
	In 3D scenes, due to light bouncing off surfaces, the incoming radiance $L_i$ can be considered as originating from another point on a different surface. This implies that $L_i$ also follows the integral form described by the rendering equation \eqref{rendering} and needs to be evaluated at the point from which the scattered light originates.  This makes the problem highly complex, and no analytical solution can be formed in general.  
 

	Path tracers solve the rendering equation via the Monte Carlo method, which draws random direction samples $\omega_i'$, and the average of samples is the estimation of the integral:
	\begin{equation}
	L(\mbx,\omega_o) \approx  L_e(\mbx) +  \frac{1}{N}\sum_{k}^{N}\frac{f_s(\mbx,\omega_{ik}',\omega_o)L_{ik}(\mbx, \omega_{ik}') |\cos\theta_{ik}|}{p(\omega_{ik}')},
	\end{equation}
	where $p(\omega_{ik}')$ is the probability density function (PDF) from which the integrator samples at point $\mbx$. Due to the complexity of light transport over the 3D space, the variance of samples is high, and many importance sampling methods have been proposed to reduce this variance (e.g., \cite{VEACH98, MNEE, ManyLights1}).
	These methods usually focus on importance sampling of single components of the rendering equation. Here, a universal importance sampling technique for the whole product of the rendering equation can help further improve sampling efficiency. Path guiding, which we describe next, is a genre of methods developed along this direction.
 
	Path guiding methods aim to find sufficient approximation of the incident radiance distribution, which serves to achieve better importance sampling that follows the global illumination distribution rather than local BSDF distribution. 
    Jensen~\shortcite{JENSEN95} and Lafortune et al.~\shortcite{Lafortune95} first proposed to fit the incident light distribution for more efficient indirect lighting sampling. 
    Vorba et al.~\shortcite{VORBA14}
    fit a Gaussian mixture to model the distribution estimated by an explicit photon tracing pass.    
    M\"{u}ller et al.~\shortcite{MUL17,PPG19}
    proposed the ``Practical Path Guiding (PPG)'' algorithm
    to model the distribution using an SD-tree approach. Based on these techniques, full product guiding methods have also been proposed. Herholz et al.~\shortcite{Herholz16} calculated an approximation of full product on top of an earlier work \cite{VORBA14}, and Diolatzis et al.~\shortcite{LTCPG} calculated an approximation on top of another study \cite{MUL17} to achieve product guiding. Both of these methods operate at the cost of higher overhead as the approximated product needs to be calculated multiple times from the corresponding learned incident radiance distribution.
    The above techniques share the same idea of partitioning the space and progressively learning discrete distributions.  Each discrete distribution is shared among points within the spatial partition.  A major issue of this approach is parallax error: the above techniques fail to accurately learn the distribution for close-distance incident radiance where the incoming light quickly changes within the same spatial partition. Ruppert et al.~\shortcite{PAPG} proposed a parallax-aware robust fitting method to address this issue with a discrete approach. We compared their approach with our network-based approach in \secref{sota}.
    Recent research shows that good path guiding requires a proper blending weight between BSDF and 
    product-driven distributions~\cite{MUL20}, and the learned distribution should be variance-aware since the samples are 
    not zero-variance \cite{Variance-awarePG}. These findings can help to achieve a more robust path guiding framework. 


Recently, neural networks have been used to model scattered radiance distribution; both online and offline learning methods have been proposed. 
Zhu et al.~\shortcite{ZHU21} used neural networks to estimate a quad-tree representation of incident radiance distribution using nearest photons as input. Currius et al.~\shortcite{EG2020} used convolutional neural networks (CNN) to estimate incident radiance represented by spherical Gaussians; this work is for real-time rendering, but the estimated radiance distribution could be used for path guiding.
Zhu et al.~\shortcite{SHANDONG21} applied an offline-trained neural network to efficiently sample complex scenes 
with lamps. This technique requires training a U-Net for more than 10 hours for just a single 
light source, while the estimated distribution is only a $16\times 16$ 2D map, which is not sufficient for representing general indirect lighting distributions. A much 
higher resolution is required for robustly guiding over the entire 3D scene; for instance, PPG's quad-tree approach can represent 
a resolution of $2^{16} \times 2^{16}$). These methods are categorized as offline learning, since they require training a network offline with massive training samples using ground truth distributions. 

Our framework, however, is categorized as online learning, which learns the distribution on the fly, without a ground truth distribution as reference.
 Previously, M\"{u}ller et al.~\shortcite{MUL20} adopted normalizing flow~\cite{Kobyzev_2021} to model the full product of incident radiance distribution. However, with an implicit density model like normalizing flow, each sample/density evaluation requires a full forward pass of multiple neural networks, which introduces heavy computation costs. In a modern path tracer with multiple importance sampling (typically, BSDF sampling and next event estimation (NEE)), this means full forward pass needs to be executed for each surface sample at least two times.
 Moreover, the training process requires dense usage of differentiable transforms, which makes the training slower than that for regular 
 neural networks. Indeed, M\"{u}ller et al.~\shortcite{MUL20} used two GPUs specifically for the normalizing flow's neural network computation along with the CPU-based path tracing implementation; nevertheless, the sampling speed was still only 1/4 of PPG. Our work, in contrast, proposes the use of an explicitly 
 parameterized density model in closed-form that can be learned using a small MLP. The neural network is used to 
 generate the closed-form distribution model, rather than actual samples; therefore, we are able to freely generate samples or 
 evaluate the density after a single forward pass. With careful implementation, we show that our method can reach a similar sampling 
 rate to that of PPG, while the result has lower variance. In research that is parallel with our work, Dong et al. ~\shortcite{PKU2023} used a small neural network to estimate per-shading-point distribution, which is similar to our approach. However, we also propose the use of NASG, a novel anisotropic model, in place of classic von Mises-Fisher distribution. NASG helps to further improve the fitting accuracy and guiding efficiency. We highlight the benefits through experiments in \secref{Eval}.

	\subsection{GPU Path Tracing}
	GPU path tracing has gained significant attention in recent years due to its high performance, supported by the GPU's ability to concurrently render many pixels. While many recent research works have focused on constructing real-time path tracers~\cite{RESTIR1,RESTIR2,RESTIR3,NRC,DENOISER}), GPU-based production renderers leverage GPU to render fully converged, noise-free results faster~\cite{REDSHIFT,OCTANE}. 
 Compared to CPU architecture, GPU requires a sophisticated programming design for better concurrency, which is needed for overcoming the large memory latency of GPU. One of the general idea of this is ray re-ordering \cite{RayReordering, RayShuffle}. Laine et al.~\shortcite{WAVEFRONT} proposed a new architecture that splits full path tracing computation into stages (small kernels) and processes threads in the same stage to maximize performance. Recently, Zheng et al.~\shortcite{luisa} proposed a more sophisticated framework for GPU path tracing, which leverages run-time compilation for high performance and flexible implementation. While several GPU-based renderer products are available and it has been proven that path guiding can improve sampling efficiency, 
few products have provided GPU-based path guiding implementation. 
This is because existing path guiding approaches are designed for CPUs, and specialization for GPUs 
only emerged recently at the research level~\cite{GPUPG1,GPUPG2}. 
Our work provides a low-cost continuous path guiding option for GPU-based path tracers.

	\subsection{Density Models} \label{sec:DensityModels}
	Parametric density models have been extensively explored in statistics. Exponential distribution is a representative family, among which Gaussian mixture is the most widely used density model. Vorba et al.~\shortcite{VORBA14} used a 2D Gaussian mixture to model incident radiance distribution; however, since the domain of a 2D Gaussian is the entire 2D plane, when mapping it to unit sphere, it is necessary to discard samples outside the domain, which affects computation efficiency. Dodik et al.~\shortcite{SDMM} further proposed using 5D Gaussians to model incident radiance distribution over the space. By using a tangent space formulation, it greatly reduced the number of discarded samples.
    
    In the context of physically-based rendering, spherical models have been widely leveraged. The spherical Gaussian (SG) is widely used for radiance representation and density modelling (e.g., \cite{SG}). Actually, a normalized SG is equivalent to the von Mises-Fisher (vMF) distribution in 3D. It has several desirable properties, such as computational efficiency and analytical tractability for integrals. However, its expressiveness is limited due to its isotropic nature, which restricts the shape of the distribution on the sphere. The Kent distribution \shortcite{Kent} presents a possible remedy by generalizing the 3D vMF model to a five-parameter anisotopic spherical exponential model, providing higher expressiveness. 
    However, its application within our framework is impeded by three significant limitations: high computational cost, the absence of a direct sampling method, and issues with numerical precision, particularly when the parameters for controlling concentration are high. These limitations are elaborated upon in \secref{kent}.
 
    Other recent works proposed spherical models specifically for graphics. Xu et al.~\shortcite{ASG} proposed an anisotropic spherical Gaussian model to achieve a larger variety of shapes; however, no analytic solution was found for its integral, which is problematic for normalization.  

    Heitz et al.~\shortcite{LTC} proposed another anisotripic model with a closed-form expression called linearly transformed cosines (LTC). However, the integral of LTC requires computation of the inverse matrix, and one component requires in total $12$ scalars to parameterize a component. 
 
	Another family of density models widely adopted in the rendering context is the polynomial family, 
	among which spherical harmonics is the most commonly used model. Polynomial models have been  
	extensively used in graphics~\cite{PRT, PolynomialRegression}. They can be easily mapped to unit spheres 
	to represent spherical distributions. However, polynomial models are generally limited to capturing low-frequency distributions. To represent high-frequency distributions, a very high degree is required, leading to a significant increase in the number of parameters. Furthermore, there is currently no efficient approach to directly sample a polynomial distribution, although there does exist some relatively expensive importance sampling schemes (e.g., \cite{ImportanceSamplingSH}).
	These were the main challenges in our pilot study that made us give up the idea of adopting them for our 
	path guiding scheme.   

	Learning-based density models have been drawing more attention recently~\cite{MUL20,MDMA}. 
	The model based on normalizing flow can successfully learn complex distributions~\cite{MUL20} 
	but suffers from the implicitness and heavy computation mentioned earlier.  
	Marginalizable Density Model Approximation (MDMA)~\cite{MDMA} has been proposed as a closed-form learning-based density 
	model, which is essentially a linear combination of multiple sets of 1D distributions.
 MDMA showed that closed-form normalization is an essential factor in learning density models, and 
	inspired by their work, we propose such a model for our application. 

	\section{Overview}
	\begin{figure}
		\centering
		\includegraphics[width=0.98\linewidth]{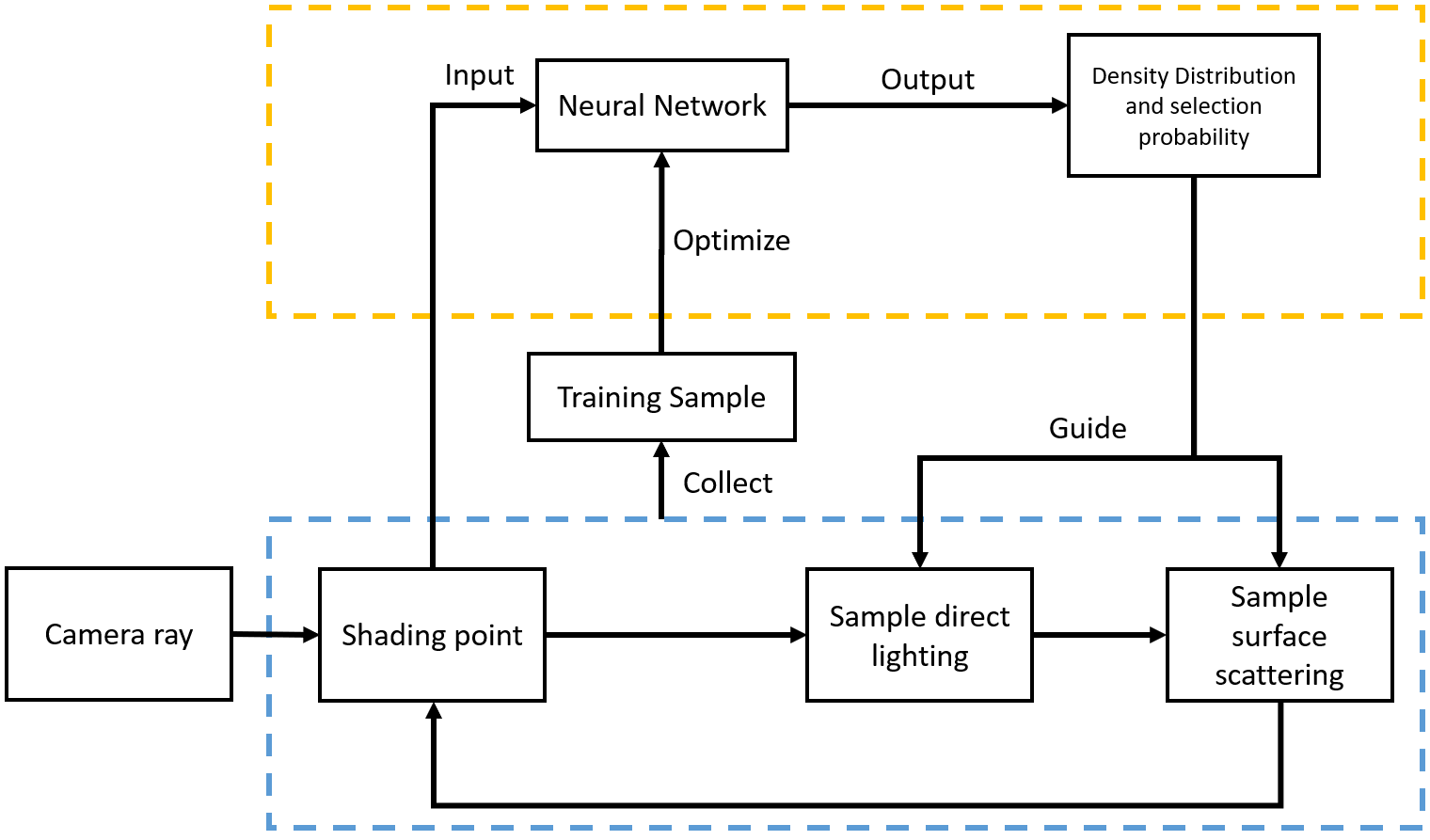}
		\caption{Overview of proposed framework. Area in blue frame is computation of a classic path tracer. Area in yellow frame shows how we learn the light distribution with a neural network and use the estimated explicit density model to guide further sampling.  }
		\label{fig:overview}
	\end{figure}
	The goal of our work is to model the distribution of the product of incident light over the unit sphere for every unique shading point $\mbx$ so that the product can be fully importance-sampled (see \secref{DDM} for  details of our density model). In other words, we map the shading point's 3D position and auxiliary data to a parameterized spherical distribution.
 
	As shown in Fig. \ref{fig:overview}, our framework learns the distribution online, in a progressive manner. In every rendering iteration, the image is rendered at one sample per pixel, and a subset of the pixel samples are collected to train an MLP (see \secref{progressive} for details of collecting strategy). After the image is rendered and training samples are collected, we train the MLP using an optimizer implemented with Libtorch (C++ frontend of Pytorch \shortcite{Pytorch}) as described in \secref{MLP}. When rendering an image,  every time a ray hits a shading point $\mbx$, our framework uses a learned MLP (whose weights are updated every $M$ iterations as described in \secref{progressive}) to estimate the product distribution, and it draws the sample from either the BSDF distribution or the learned one, based on the learned selection probability.
 
	
	\section{Normalized Anisotropic Spherical Gaussian Mixture}
	\label{sec:DDM}
	In this section, we describe our proposed density model we call  
	the Normalized Anisotropic Spherical Gaussian mixture (NASG), which allows us to efficiently learn the  
	light distribution at each shading point. 
	We first review the existing density models that inspired our    
	model in \secref{background}. We then describe our model itself in \secref{NASGM}. 
	
	
	\subsection{Background}
	\label{sec:background}
	One of the main requirements for progressive learning of a distribution with neural networks is that models be easily normalizable, implying that a model must have a closed-form integral.
	
	\paragraph{Marginalizable Density Model Approximation}
	\citeN{MDMA} proposed Marginable Density Model Approximation (MDMA), which can be effectively optimized via deep learning. A bivariate dimensional distribution can be modeled as
	\begin{equation}
	D(x,y) = \sum_{i,j}A_{ij}\phi_{1i}(x)\phi_{2j}(y),
	\end{equation}
	where $\phi_{1i}$ and $\phi_{2j}$ are 1D normalized distributions and $A_{ij}$ are normalized coefficients that sum to $1$.
	
	In practice, the 1D distributions in MDMA can be piece-wise linear or piece-wise quadratic spline models; however, with its limited number of components, MDMA can learn only a coarse distribution. Therefore, it always gives poor results for distributions with high-frequency spots. Despite the limited accuracy, MDMA works well in learning distributions from sparse noisy samples, compared with non-normalizable models (\eg polynomial model, spherical harmonics). During training, samples with high energy can lower the contribution of other areas, and eventually the training converges. This feature is crucial in learning distributions from sparse training samples.
	\paragraph{Normalized Gaussian Mixtures}
	Normalized Gaussian mixtures can address the accuracy limitations of MDMA:
	\begin{equation}
	D(\boldsymbol{\mbx}) = \sum_{i}^{N} A_i\frac{G_i(\boldsymbol{\mbx};\theta_i)}{K_i},
	\end{equation}
	where $G_i(\boldsymbol{\mbx};\theta_i)$ are Gaussian distributions parameterized by $\theta_i$, $K_i$ are normalizing factors, and $A_i$ are normalized weights of Gaussians such that $\sum_{i}A_i = 1$. 
	Gaussian mixtures are highly expressive and also easily normalizable: we look into two common Gaussian distributions that can represent the spherical distributions needed for modeling the lighting: spherical Gaussian and anisotropic spherical Gaussian. 
	\paragraph{Spherical Gaussian}
	Spherical Gaussian (SG) is a variant of the univariate Gaussian function defined in the spherical domain:
	\begin{equation}
	G(\mbv;\boldsymbol{\mu},\lambda) = \exp(\lambda (\boldsymbol{\mu} \cdot \mbv - 1)),
	\end{equation}
	where $\mbv$ is a unit vector specifying the direction, $\boldsymbol{\mu}$ is the lobe axis, and $\lambda$ is a parameter that controls the ``sharpness'' of the distribution. The normalizing term of an SG can be computed by 
	\begin{equation}\label{eq:SG integral}
	K = \int_{S^2} G(\mbv;\boldsymbol{\mu},\lambda) d\omega = \frac{2\pi}{\lambda}(1-e^{-2\lambda}).
	\end{equation}
	Normalized SG is equivalent to the vMF distribution in 3D. Since SG is defined in the spherical domain and is suitable for representing all-frequency light distribution at each sample point, mixture models based on SG have been applied in representing the light map used for precomputed radiance transfer (e.g., \cite{SG, EG2020}).     
	However, SG is an isotropic univariate model, and thus less expressive; for example, it cannot represent anisotropic distributions.  
	
	\paragraph{Anisotropic Spherical Gaussian}
 To model anisotropic distributions, Xu et al.~\shortcite{ASG} proposed Anisotropic Spherical Gaussian (ASG) to model the light map.  ASG can be written as   
 \begin{equation}
	 G(\mbv;[\boldsymbol{x},\boldsymbol{y},\boldsymbol{z}],[a,b],c) = c\cdot S(\mbv; \boldsymbol{z}) \cdot e^{(-a (\mbv \cdot \boldsymbol{x})^2 - b(\mbv \cdot \boldsymbol{y})^2)},
\end{equation}
where $\boldsymbol{z}$, $\boldsymbol{x}$, and $\boldsymbol{y}$ are the lobe, tangent, and bi-tangent axes, respectively, and $[\boldsymbol{x}, \boldsymbol{y}, \boldsymbol{z}]$ form an orthonormal frame; $a$ and $b$ are the bandwidths for the $\boldsymbol{x}$- and $\boldsymbol{y}$-axes, respectively, and  $c$ is the lobe amplitude.  Xu et al.~\shortcite{ASG} successfully modeled complex lighting conditions and  rendered anisotropic metal dishes using ASG. 
On the other hand, ASG does not have a closed-form solution for the integral, which inhibits its usage for our purpose.  

 \paragraph{Kent Distribution} \label{sec:kent}
 Another candidate is the Kent distribution~\cite{Kent},
which is also an anisotropic spherical distribution:
 \begin{equation}
    K(\mbv; \kappa, \beta) = \frac{1}{C(\kappa, \beta)} \exp\left( \kappa(\mbv \cdot \mbz) + \beta ((\mbv \cdot \mbx)^2-(\mbv \cdot \mby)^2) \right),
 \end{equation}
 where $\kappa, \beta$ 
are parameters that control anisotropic concentration with $0<2\beta<\kappa$,
and $[\boldsymbol{x}, \boldsymbol{y}, \boldsymbol{z}]$ form an orthonormal frame in the same way as ASG.  The Kent distribution has a closed-form integral, i.e., its normalizing constant:
\begin{equation}
C(\kappa, \beta) = 2\pi \sum_{i=0}^{\infty} \frac{\Gamma(i+0.5)}{\Gamma(i+1)} \beta^{2i} (\frac{\kappa}{2})^{-2i-\frac{1}{2}}I_{2i+\frac{1}{2}}(\kappa),
\end{equation}
where
$\Gamma(\cdot)$ is the gamma function, and $I_n(\kappa)$ is the modified Bessel function of the first kind. However, since this integral entails the evaluation of a nested infinite series (since $I_n(\kappa)$ is also an infinite series), sufficient precision necessitates the computation of a significant number of terms, thus imposing a substantial computational overhead. In the context of computer science, the precision of the Kent distribution is further constrained by the limitations on floating-point number precision. Specifically, when the value of $\kappa$ is large (e.g., $\kappa > 20$), both the non-normalized density and the normalizing constant approach the representational boundary of floating-point numbers, resulting in a lack of precision. This leads to either a biased result or disrupted learning (where an NaN loss is generated and corrupts the learning process). Within the realm of Monte Carlo rendering, an additional limitation emerges, since Kent distribution lacks a direct sampling method, which necessitates the use of a more computationally expensive rejection method.

	\subsection{Normalized Anisotropic Spherical Gaussian}
	\label{sec:NASGM}
\begin{figure}[t]
		\begin{tabular}{cc}
			\includegraphics[trim = 40 40 40 40, clip,height=3.2cm]{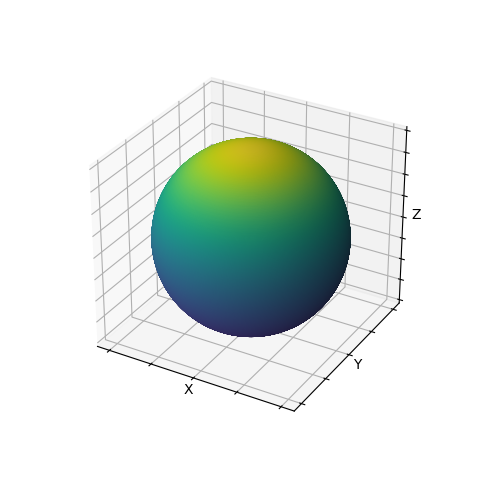}&
			\includegraphics[trim = 40 40 40 40, clip,height=3.2cm]{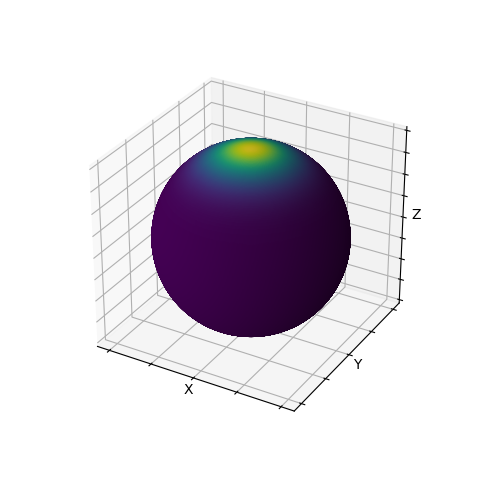}\\
			$\lambda = 1, a = 0$&$\lambda = 10, a = 0$\\
			\includegraphics[trim = 40 40 40 40, clip,height=3.2cm]{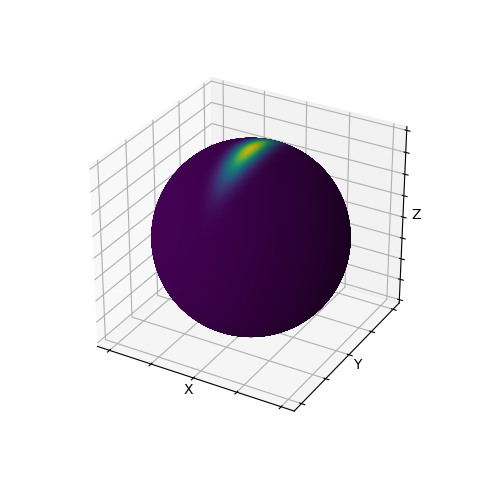}&
			\includegraphics[trim = 40 40 40 40, clip,height=3.2cm]{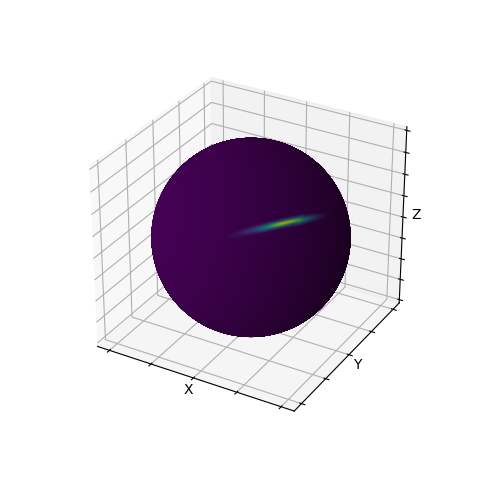}\\
			$\lambda = 10, a = 10$&$\lambda = 25, a = 100$, rotated
		\end{tabular}
		\caption{Visualization of NASG component $G$ with different parameters. Note that $G$ agrees with spherical Gaussian when $a = 0$. }
		\label{fig:nasg}
	\end{figure}

	Inspired by the models in \secref{background}, we believe a spherical anisotropic exponential distribution can help to achieve more accurate results in our framework. While the Kent distribution exists as a potential candidate, its inherent limitations - namely, the necessity of approximating its integral, the precision issue, and the absence of a direct sampling algorithm - prompt us to develop our own distribution, which we call the Normalized Anisotropic Spherical Gaussian:
	
	\begin{equation}
	\begin{aligned}
	&G(\mbv ;[\boldsymbol{x,y,z}], \lambda, a) \\
	&\quad= \begin{cases}\exp \left(2 \lambda\left(\frac{\mbv \cdot \boldsymbol{z}+1}{2}\right)^{1+\frac{a(\mbv \cdot \boldsymbol{x})^2}{1-\left(\mbv \cdot \boldsymbol{z}\right)^2}}-2 \lambda\right)\left(\frac{\mbv \cdot \boldsymbol{z}+1}{2}\right)^{\frac{a(\mbv \cdot \boldsymbol{x})^2}{1-(\mbv \cdot \boldsymbol{z})^2}} & \text { if } \mbv \neq \pm\boldsymbol{z} \\
	1 & \text { if } \mbv = \boldsymbol{z} \\
	0 & \text { if } \mbv = -\boldsymbol{z}, \end{cases}
	\end{aligned}
	\end{equation}
 where $[{\boldsymbol{x},\boldsymbol{y},\boldsymbol{z}}]$ forms an orthogonal frame as above, $\lambda$ is  sharpness, and $a$ controls the eccentricity.
	 \figref{nasg} shows examples of the NASG component with different parameters.

The derivation of NASG proceeds as follows. First, we employ the standard spherical coordinate expression of $\mbv\ne\pm\boldsymbol{z}$ in terms of the polar angle $\theta$ and the azimuthal angle $\phi$ (with respect to the orthonormal frame $[{\boldsymbol{x},\boldsymbol{y},\boldsymbol{z}}]$) to obtain the equation
\begin{equation}\label{eq:Jacobian}
    \left(\frac{\mbv \cdot \boldsymbol{z}+1}{2}\right)^{\frac{a(\mbv \cdot \boldsymbol{x})^2}{1-(\mbv \cdot \boldsymbol{z})^2}}=\left(\frac{\cos\theta+1}{2}\right)^{a\cos^2\phi},
\end{equation}
which introduces the anisotropic nature of the distribution through the exponent $a\cos^2\phi$. Next, we effect a change of variables in the integral of SG (cf.~\eqref{SG integral}) so that the Jacobian in \eqref{Jacobian} emerges as part of the corresponding Jacobian. In this way, we arrive at NASG, which satisfies all requirements as a model for learning the distribution needed for unbiased rendering.

	Different from ASG \cite{ASG}, NASG has 
	an {\bf analytical closed-form solution} for its integral (see \secref{integral}),
	which makes it easy to normalize:
	\begin{equation}\label{eq:K for NASG}
	K = \int_{S^2} G(\boldsymbol{\mbv};[\boldsymbol{x,y,z}], \lambda, a) d \omega=\frac{2 \pi\left(1-e^{-2 \lambda}\right)}{\lambda \sqrt{1+a}}.
	\end{equation}
	NASG is also {\bf expressive}, since it can model anisotropic distributions, as well as {\bf numerically stable} to represent high- and low-frequency distributions accurately
	\footnote{\label{footnote:NASG} NASG is not continuous at $\boldsymbol{\mbv}=-\boldsymbol{z}$ when $a> 0$ in this form.
 Indeed, it approaches $0$ if we let $\mbv$ tend to $-\boldsymbol{z}$ along any meridian, except for the ones passing through $\pm\boldsymbol{y}$ (i.e., $\phi=\pi/2$ and $\phi=3\pi/2$), in which case NASG approaches $\exp(-2\lambda)>0$.
 However, this discontinuity does not affect our application, and it can be resolved easily by introducing an auxiliary parameter (see \secref{continuity} for details)}. 
	It is also {\bf compact} because
	it is parameterized by only $7$ scalars, making it highly efficient in GPU-based computation due to its low bandwidth requirement. Additionally, the sampling algorithm of NASG is remarkably {\bf efficient} (see \secref{sampling}). These characteristics make NASG a highly feasible and practical option for our framework.
	
	\section{Online learning of Density Model} \label{sec:MLP}
	We now describe our neural network structure for learning the parameters of our NASG model, as well as the training scheme. 
	\subsection{Network Architecture}
	\label{sec:Output}
	We intend our neural network to be as simple as possible, considering the performance requirements for practical use. 
	In this work, we use a $4$-layer MLP, where each layer has 128 units \textit{without} bias.
	
	\paragraph{Network Inputs}The input to the model consists of the location of the shading point $\mbx$, outgoing ray direction $\omega_o$, and surface normal $\mbn$. 
	The location of the shading point is first encoded into a $57$ dimensional vector by positional encoding.  Thus, the input size is 63 (57 + 3 + 3), 
	but is padded to 64 for hardware acceleration purposes\footnote{By setting padded values to 1 we achieve an alternative to the bias of hidden layers.}. We adopt the simple one-blob encoding \cite{MUL19} 
	instead of complex encoding mechanics, such as \textit{Multiresolution Hash Encoding} \cite{INP}, because there was very little improvement in our pilot study to justify the overhead such mechanics introduce. Details of the encoding scheme can be found in \tabref{encoding}.
	
	\begin{table}[h]
		\caption{Encoding scheme of network inputs}
		\label{tab:encoding}
		\begin{tabular}{ccc}
			\hline
			Parameter&Symbol&Encoding\\\hline
			Position&$p\in{\mathbb{R}^3}$&$ob(p)$\\
			Outgoing ray direction&$\omega_o\in{[-1,1]^3}$&$\omega_o$\\
			Surface normal&$\mbn\in{[-1,1]^3}$&$\mbn$
		\end{tabular}
	\end{table}
	
	\paragraph{Parameterization of NASG}
	To improve the efficiency of learning, we reduce the number of parameters used to represent the NASG model. First, we introduce $\theta,\phi,\tau$, the Euler angles representing the 
	orientation of the orthogonal basis with respect to the global axes. Accordingly, $\mbz$ and $\mbx$ of the orthogonal basis can be represented by  
	\begin{equation}
		\mbz =
	\begin{pmatrix}
		 \cos \phi \sin \theta \\
		 \sin \phi \sin \theta \\ 
		 \cos \theta
	\end{pmatrix}, \
		\mbx =
	\begin{pmatrix}
		\cos\theta \cos\phi \cos\tau - \sin\phi \sin\tau \\
		\cos\theta \sin\phi \cos\tau + \cos\phi \sin\tau \\
		-\sin\theta \cos\tau 
	\end{pmatrix}. 
	\end{equation}
	Thus, we represent NASG with the following 
	seven parameters: $cos\theta$, $sin\phi$, $cos\phi$, $sin\tau$, $cos\tau$, and $\lambda$, $a$.   
	
	\paragraph{Network Outputs} The output of our neural network is a $d$ dimensional vector $\boldsymbol{o}$, where each component corresponds to the  
 	NASG parameter at the shading point $\mbx$ and a selection probability $c$. Assuming the number of mixture components is $N$, 
	$d = 5 \times N + 2 \times N + N + 1 = 8N + 1$ (see \tabref{decoding}). 
	These parameters are further decoded to represent the actual parameters as summarized in \tabref{decoding}.
	\begin{table}[h]
		\caption{Decoding scheme for NASGM with N components}
		\label{tab:decoding}
		\begin{tabular}{ccc}
			\hline
			Parameter&Symbol&Decoding\\\hline
			$\cos\theta$, $\sin\phi$, $\cos\phi$, $\sin\tau$, $\cos\tau$&$s_0\in{R^5} \times N$&$sigmoid(s_0) \times 2 - 1$\\
			$\lambda, a$&$s_1\in{R^2} \times N$&$e^{s_1}$\\
			Component weights&$\mbA\in{R^N}$&$softmax(A)$\\
			Selection probability&$c\in{R}$&$sigmoid(c)$\\
		\end{tabular}
	\end{table}
	The mixture weights of NASG $\mbA$ are normalized with a softmax function.
	\subsection{Training}
\begin{figure}
		\centering
		\includegraphics[width=0.99\linewidth]{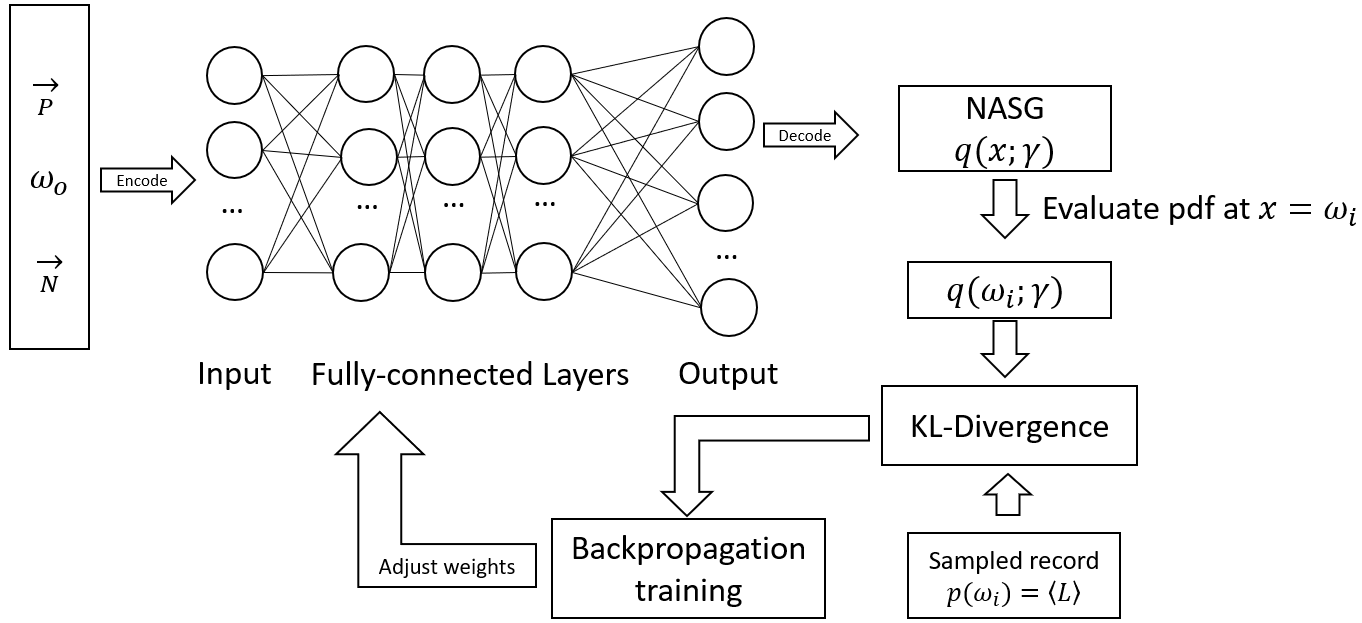}
		\caption{Training process of proposed network. We heavily apply automatic differentiation to adjust network weights  based on estimated distribution and rendering estimation $\left< L\right>$ from the sampled scattering direction $\omega_i$.}
		\label{fig:networkgraph}
	\end{figure}

	We now describe our loss function and the training process. 
	As shown in \figref{networkgraph}, we utilize automatic differentiation to train the network effectively with sparse online rendering samples. 
	Since the training data are noisy online samples, we need a loss function that is more robust than regular L1 or L2 losses.
	\paragraph{Kullback-Leibler Divergence}
	The design of a loss function for distribution learning in PBR context has been studied in several previous works \cite{MUL19, MUL20, ZHU21, SHANDONG21}. By  using $q(\omega_i;\gamma)$ to denote the NASG distribution (with NASG parameters $\gamma$) estimated by the neural network, to achieve importance sampling of the rendering equation, the optimal distribution should be proportional to the product:
 
 \begin{equation}
	q(\omega_i;\gamma) \propto f_s(\mbx,\omega_i, \omega_o)L_i(\mbx, \omega_i) |\cos\theta_i|.
 \end{equation}
	Therefore, we use 
 $p(\omega_i) = F f_s(\mbx,\omega_i, \omega_o)L_i(\mbx, \omega_i) |\cos\theta_i|$
 as our target distribution, where $F$ is a normalizing term whose value is unknown. We can still train the system without knowing $F$ as shown below. 
     
 In our framework, we use the Kullback-Leibler Divergence (KL-Divergence) to represent the likelihood between our estimated distribution $q(\omega_i;\gamma)$ and the target one:
	\begin{equation}
	D_{KL}(p(\omega_i)||q(\omega_i;\gamma)) = \int_{S^2}p(\omega_i)(\log[p(\omega_i)] - \log[q(\omega_i;\gamma)])d\omega_i.
	\end{equation}
Since $p(\omega_i)$ is $\gamma$-independent, we get 
	\begin{equation}
	\nabla_\gamma D_{KL}(p(\omega_i)||q(\omega_i;\gamma)) = -\int_{S^2}p(\omega_i)\nabla_{\gamma}\log[q(\omega_i;\gamma)] d\omega_i.
	\end{equation}
Then, the integral can be replaced with our one-sample estimation:
	\begin{equation}
	\nabla_{\gamma} D_{KL}(p(\omega_i)||q(\omega_i;\gamma)) = -\, \mathbb{E} \left[ \frac{p(\omega_i)}{\hat{q}(\omega_i;\gamma)}\nabla_{\gamma}\log[q(\omega_i;\gamma)]\right],
	\end{equation}
 where $\hat{q}(\omega_i;\gamma)$ denotes the PDF of the distribution the sample obeys during rendering (see below).
Although the right-hand side still involves the global scaling factor $F$, it
is canceled in moment-based optimizers such as Adam \cite{Adam}; accordingly, we can train the system even though there is an unknown scaling factor in $p(\omega_i)$~\cite{MUL19}.

	\paragraph{Selection Probability}
	   Following \cite{MUL20}, we do not directly use a distribution estimated by our neural network. Instead, we regard our scattering sampling process as an MIS process that blends the learned distribution and the BSDF distribution $p_{f_s}(\omega_i)$. As we also learn a selection probability $c$, as described in \secref{Output}, the MIS PDF at shading point $\boldsymbol{\mbx}$ is $\hat{q}(\omega_i;\gamma) = cq(\omega_i;\gamma) + (1-c)p_{f_s}(\omega_i)$. However, optimizing $\hat{q}(\omega_i;\gamma)$ naively leads to falling into local optima with degenerate selection probability; therefore, our final loss function blends $\hat{q}(\omega_i; \gamma)$ with $q(\omega_i;\gamma)$:
	\begin{equation}
	loss = e D_{KL}(p(\omega_i)||\hat{q}(\omega_i; \gamma)) + (1-e)D_{KL}(p(\omega_i)||q(\omega_i; \gamma)),
	\end{equation}
	where $e$ is a fixed fraction that we set to $0.2$.

	\section{Implementation Details}
    We integrate our framework into Clight, our in-house wavefront GPU path tracer for production \cite{CLIGHT}. In this section we discuss implementation details as well as design considerations.
	\paragraph{Progressive Learning and Sampling} \label{sec:progressive}
	Our framework collects samples to train the network online, and we use the network to generate a distribution that is then blended with the BSDF sampling distribution using the learned selection probability $c$. We further multiply an extra coefficient $b$ to compute an actual blending weight $c'=bc$, where $b$ is initially set to 0 and gradually increased to 1, such that the framework initially relies on the BSDF distribution but gradually switches to the learned distribution $\hat{q}(\omega_i;\gamma)$. We use a fixed-step strategy to ensure that our guided sampling uses a sufficient number of samples for learning the distribution: for every $M$ images rendered, we increase $b$ by $\frac{1}{B}$. In our implementation, we set $M=4$ and $B=64$. This could require adjustments according to the rendering task, and further discussions are given in \secref{Discussion}.

    For rendering images while progressively learning the distribution, previous works~\cite{MUL17, MUL19, MUL20} blended the rendered image samples 
    based on their inverse variance.  Such a process  introduces extra memory and computation overhead, especially for a renderer that produces arbitrary output variables (AOVs). Based on the observation that the variance of individual image samples decreases as the learning proceeds, we use a simplified method that scales the weights of accumulated results according to the training steps. This essentially gives more weight to later samples in a progressive manner. The weight scaling process stops when the training step count reaches $M \times R$. Since the rendering result at each step can be seen as an unbiased estimation of the ground truth, and the weighted average can be seen as an MIS process with different sampling techniques (in this case, different distributions), the result remains unbiased.

	\paragraph{Network Optimization}
The process of distribution learning runs on GPU after every rendering iteration.
  We use Libtorch, the C++ frontend of Pytorch \cite{Pytorch}, for the implementation, along with the Adam optimizer with its learning rate set to 0.002. 
 When rendering an image, we split the image into smaller tiles of size $l \times l$ ($l$ is a real number, and $l \geq 1$) and collect data from one pixel per tile in each render iteration. After the iteration of rendering, we update $l$ based on collected sample size $s$ and maximum size $S$ as $l = \max(1, l\sqrt{\frac{s}{S}})$; consequently, after rendering the full image plane, we can get an $s$ close to $S$. Note that if $s$ exceeds $S$, we only record and keep $S$ samples for training.
 
 After each render iteration, the collected data are directly used for progressively training the model, and the training step is determined by $T = \nu \cdot \lceil \frac{S}{t} \rceil$ so that every sample will be trained at least $\nu$ times. Larger $S$ and $\nu$ help to improve the learning accuracy, but the training overhead introduces a significant trade-off.  Here, we set the following parameters: maximum size of training samples in each iteration $S = 2^{16}$, training batch size $t = 2^{12}$, and step factor $\nu = 1$. The differences in the results obtained with different configurations are revealed by experiment in \secref{Performance}.

	\paragraph{Direct Lighting with NEE}
	Our implementation uses MIS to blend guided scattering directions with Next Event Estimation (NEE) to sample direct lighting. NEE helps to improve the quality of the images, especially at the beginning of the training stage. When recording radiance to a training sample, we use the MIS-weighted value. It should be noted that MIS is only practical because our model is an explicit model that allows us to freely sample the distribution and evaluate PDF in a certain direction, which could be computationally expensive for implicit models such as ~\cite{MUL19,MUL20}.

	\paragraph{Wavefront Architecture}
	Neural network computation can be greatly accelerated with hardware (\eg Tensor Core). However, such hardware is usually designed for batched execution, in which a single matrix multiplication involves multiple threads. This conflicts with the classic thread-independent ``Mega-kernel'' path tracing implementation. Our implementation adopts ``Wavefront path tracing'' architecture \cite{WAVEFRONT} to solve this problem. In a wavefront path tracer, all of the pixels at the same stage are executed concurrently. Under this condition, we are able to ``insert'' our neural network calculation seamlessly right before the sampling/shading stage and minimize the computation overhead for best performance. The forward pass of our MLP does not require gradient calculation, so we implement it on the GPU with CUBLAS, and the resulting implementation is substantially faster than when using existing deep learning frameworks. Moreover, this implementation achieves a low overhead that makes our framework practical with conventional hardware, the performance of which we evaluate in \secref{Performance}.
	
    \section{Evaluation} \label{sec:Eval}

    We evaluated our framework in three ways. First, we integrated our framework into our in-house GPU production renderer. We evaluated its variance reduction efficiency by comparing it to that of plain path tracing. To demonstrate the benefits of our NASG and product guiding, we compared the results rendered with incident radiance guiding and other density models. Second, we evaluated its raw performance through a comparison with the results 
    produced by path tracing in the same amount of time; here, our method was executed multiple times under different meta-parameter configurations. 
    Finally, we implemented our framework in Mitsuba and compared the results 
    produced with the same number of samples
    by several state-of-the-art methods, including Practical Path Guiding (PPG, \cite{PPG19}), Robust Fitting of Parallax-Aware Mixtures for Path Guiding (PAVMM, \cite{PAPG}), and Neural Importance Sampling (NIS, \cite{MUL19}). We used mean absolute percentage error (MAPE) for quality measurement. MAPE is calculated by dividing the absolute error by the value of the ground truth. We added a small $\epsilon=0.01$ to the denominator to avoid divisions by zero.
    When calculating the MAPE from all of the rendering results,  we discarded 0.1\% of the pixels with the highest error before computing the average.

    \subsection{Variance Reduction}
    We implemented our framework on top of our in-house GPU production renderer and evaluated the quality of the images by comparing them with those produced by plain path tracing using the same number of samples per pixel (SPP). We implemented both the sampling algorithm and the optimizer on the GPU, leaving the CPU to only perform scheduling tasks. We also compared different guiding schemes, i.e., guiding with incident radiance distribution, cosine-weighted radiance distribution, and full product distribution. In addition, we compared NASG with SG, MDMA, 2D Gaussians, and Kent distribution. Here, we used 8-component NASG (64 parameters). For other density models, we adjusted the number of parameters of each Gaussian to match that of NASG: SG uses 14 components (70 parameters), 2D Gaussians uses 12 components (72 parameters), Kent distribution uses 8 components (64 parameters), and MDMA uses 8 1D distributions for both dimensions (80 parameters). For Kent distribution, even when calculations are performed in the logarithmic space to mitigate the precision issues, we found it necessary to restrict the range of $\kappa$ to $0 < \kappa < 20$, and accordingly constrained $0 < 2\beta < \kappa$ to prevent biased results or unstable learning.
    
\begin{figure*}[h]
        \begin{tabular}{cccc}
            \includegraphics[width=0.2\linewidth]{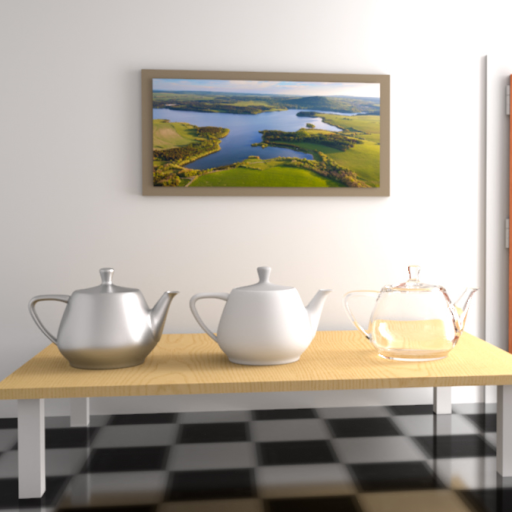} &\includegraphics[width=0.2\linewidth]{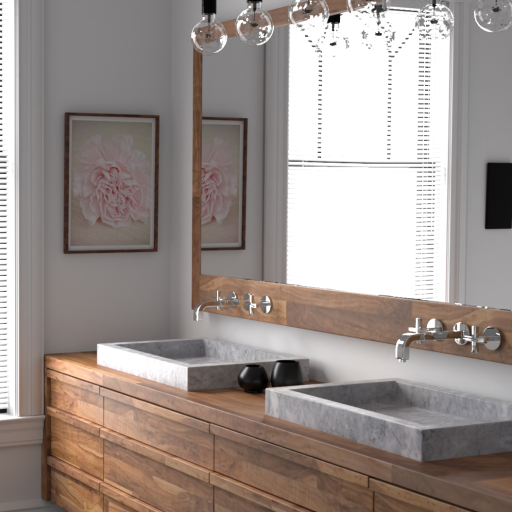} 
             \includegraphics[width=0.2\linewidth]{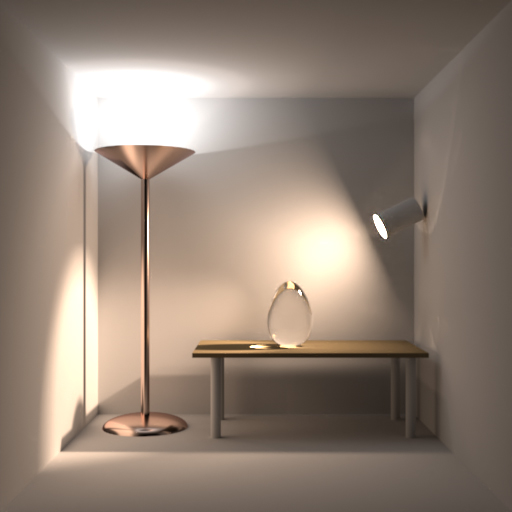} &\includegraphics[width=0.2\linewidth]{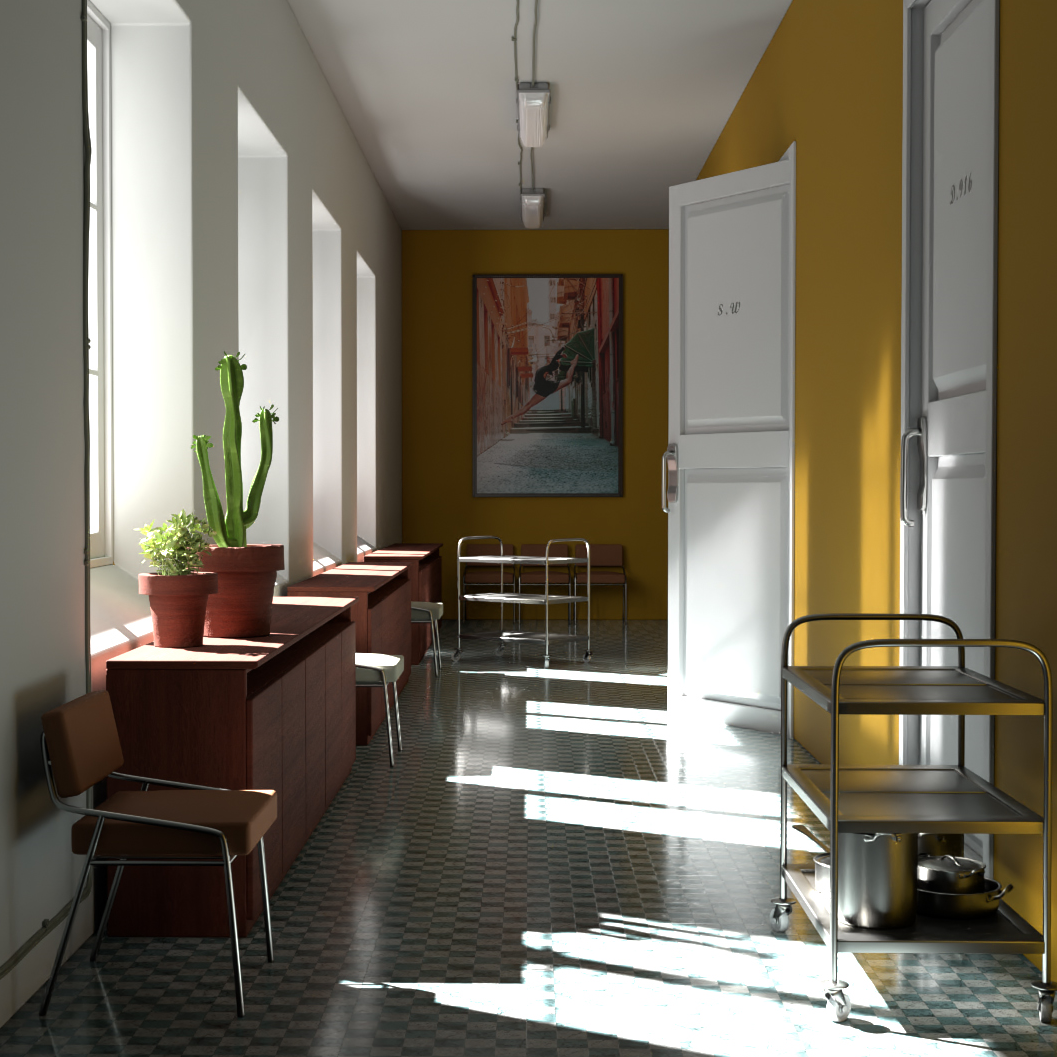}  \\ \includegraphics[width=0.2\linewidth]{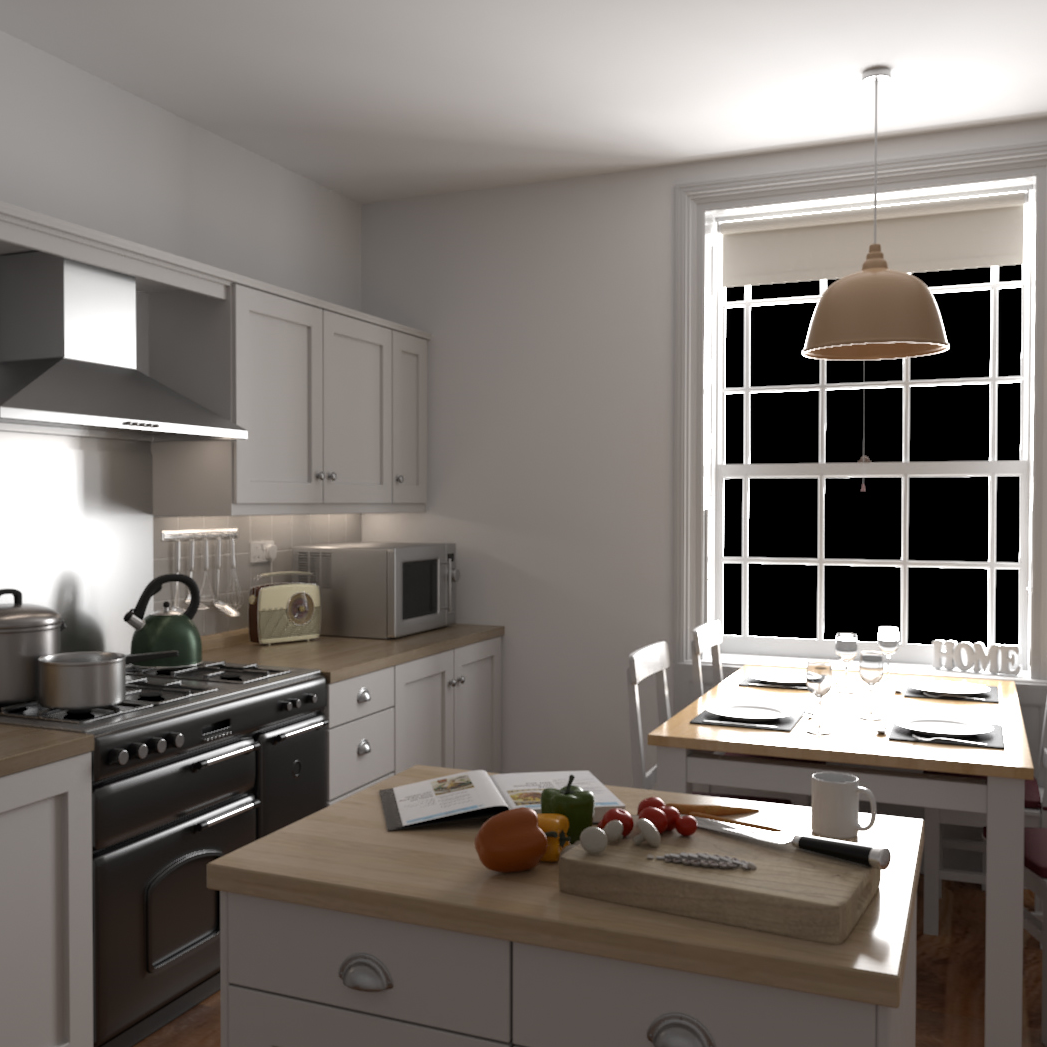} &\includegraphics[width=0.2\linewidth]{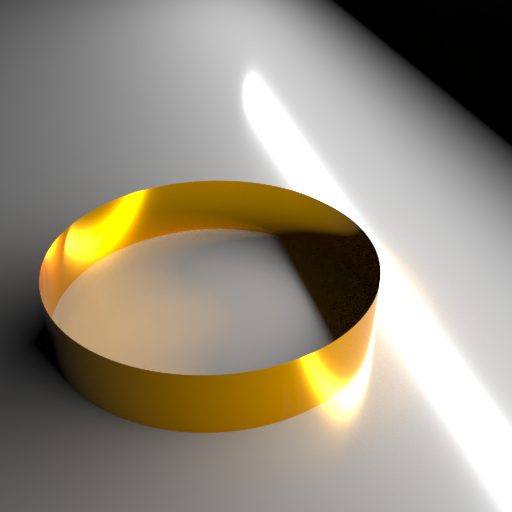}  \includegraphics[width=0.2\linewidth]{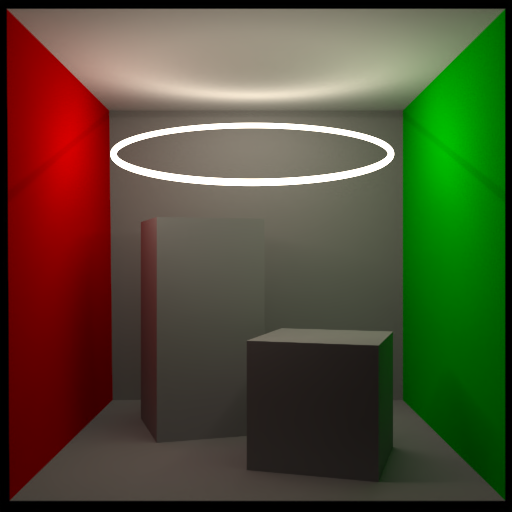} &\includegraphics[width=0.2\linewidth]{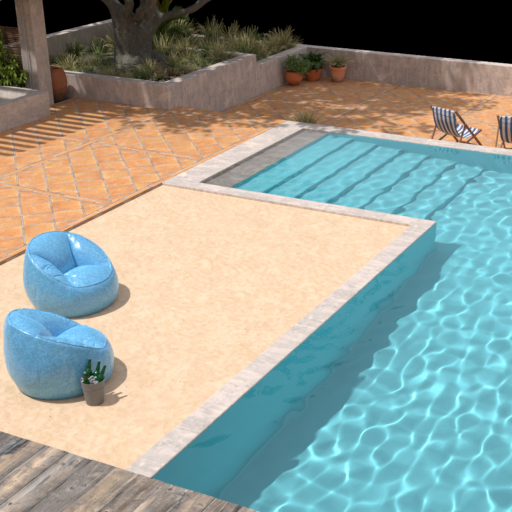}  \\ 
        \end{tabular}
        \caption{Eight scenes used to evaluate proposed GPU implementation. From left to right: \textsc{Ajar, Bathroom, Bidir, Corridor, Kitchen, Glossy Tube, Box, Pool}.}
        \label{fig:preview_gpu}
    \end{figure*}

    We run our GPU path tracer on a conventional PC with a 4-core i7 9700K CPU and an Nvidia 2080ti GPU. The implementation of our framework introduces roughly 600 MB memory overhead on the GPU, mainly for buffering batched network execution. Our Pytorch optimizer takes another 600 MB of GPU memory to cache tensors and gradients. The test 3D scenes are shown in \figref{preview_gpu}, and we render the scenes with NEE enabled. All of the rendered images are included in the supplemental material. 
    \tabref{variance_reduction} shows the results of the quantitative evaluation. 

    \begin{table*}[h]
    \caption{Errors of the results from variance reduction experiment, reported in MAPE. We rendered a variety of 3D scenes at 1024 samples per pixel (SPP), guided with our framework and different learned distributions. Scenes were rendered in different resolutions for a larger variety. We compare the learning distribution of incident radiance using NASG (NASG Li) with both the cosine-weighted incident radiance (+cos$\theta$) and the full product distribution. We also compare the learning distribution of full product with different density models, i.e., spherical Gaussians (SG), MDMA (MDMA), 2D Gaussians (G2D), and Kent distribution (KENT). The time cost of each method is provided for completeness.} 
		\label{tab:variance_reduction}
\centering
\begin{tabular}{lrrrrrrrrr}
\toprule
Scene && NASG Li & +cos$\theta$ & Full Product & SG & MDMA & G2D &KENT& PT \\
\midrule
\multirow{2}{*}{\textsc{Ajar}} & MAPE & 0.293 & 0.290 &\textbf{0.045} & 0.055 & 0.084 & 0.071&0.066&0.154 \\
 & Time (s) & 250 & 259 & 264 & 249 & 325 & 208&1641&65 \\
\midrule
\multirow{2}{*}{\textsc{Bathroom}} & MAPE & 0.085 & 0.073 &\textbf{0.064} & 0.066 & 0.081 & 0.078&0.098&0.140 \\
 & Time (s) & 360 & 373 & 381 & 358 & 469 & 300&2261&170 \\
\midrule
\multirow{2}{*}{\textsc{Bidir}} & MAPE & 0.086 & 0.079 &\textbf{0.058} & 0.066 & 0.133 & 0.090 &0.137& 0.319 \\
 & Time (s) & 115 & 120 & 126 & 119 & 151 & 100&793&25 \\
\midrule
\multirow{2}{*}{\textsc{Box}} & MAPE & 0.056 & 0.056 & \textbf{0.051} & 0.061 & 0.081 & 0.076 &0.066& 0.118 \\
 &Time (s) & 83 & 86 & 87 & 83 & 108 & 69&545&17 \\
\midrule
\multirow{2}{*}{\textsc{Corridor}} & MAPE & 0.148 & 0.142 & \textbf{0.114} & 0.117 & 0.150 & 0.132 & 0.146&0.243 \\
 & Time (s) & 209 & 211 & 220 & 212 & 272 & 178&1268&83 \\
\midrule
\multirow{2}{*}{\textsc{Kitchen}} & MAPE & 0.102 & 0.092 & \textbf{0.042} & 0.043 & 0.051 & 0.072 & 0.046&0.078 \\
 & Time (s) & 303 & 308 & 309 & 302 & 395 & 253&1980&126 \\
\midrule
\multirow{2}{*}{\textsc{Glossy Tube}} & MAPE & 0.010 & 0.011 & \textbf{0.009} & 0.015 & 0.010 & 0.015 &0.010& 0.018 \\
 & Time (s) & 134 & 136 & 136 & 133 & 174 & 111&877&61 \\
\midrule
\multirow{2}{*}{\textsc{Pool}} & MAPE & 0.056 & 0.041 & \textbf{0.041} & 0.051 & 0.209 & 0.230 &0.287& 0.321 \\
 & Time (s) & 300 & 310 & 318 & 299 & 391 & 250&1969&156 \\
\bottomrule
\end{tabular}
\end{table*}

\begin{figure*}[h]
		\begin{tabular}{@{\hspace{5pt}}r@{\hspace{5pt}}r@{\hspace{5pt}}c@{\hspace{5pt}}c@{\hspace{5pt}}c@{\hspace{5pt}}c@{\hspace{5pt}}c@{\hspace{5pt}}c@{\hspace{5pt}}c@{\hspace{5pt}}}
			&&NASG Li&+cos$\theta$&Full Product&SG&MDMA&G2D&KENT
			\\
			\rotatebox[origin=c]{90}{(a) \textsc{Corridor}}&
			\begin{minipage}{0.3\linewidth}
                \centering
				\includegraphics[height=3.0cm]{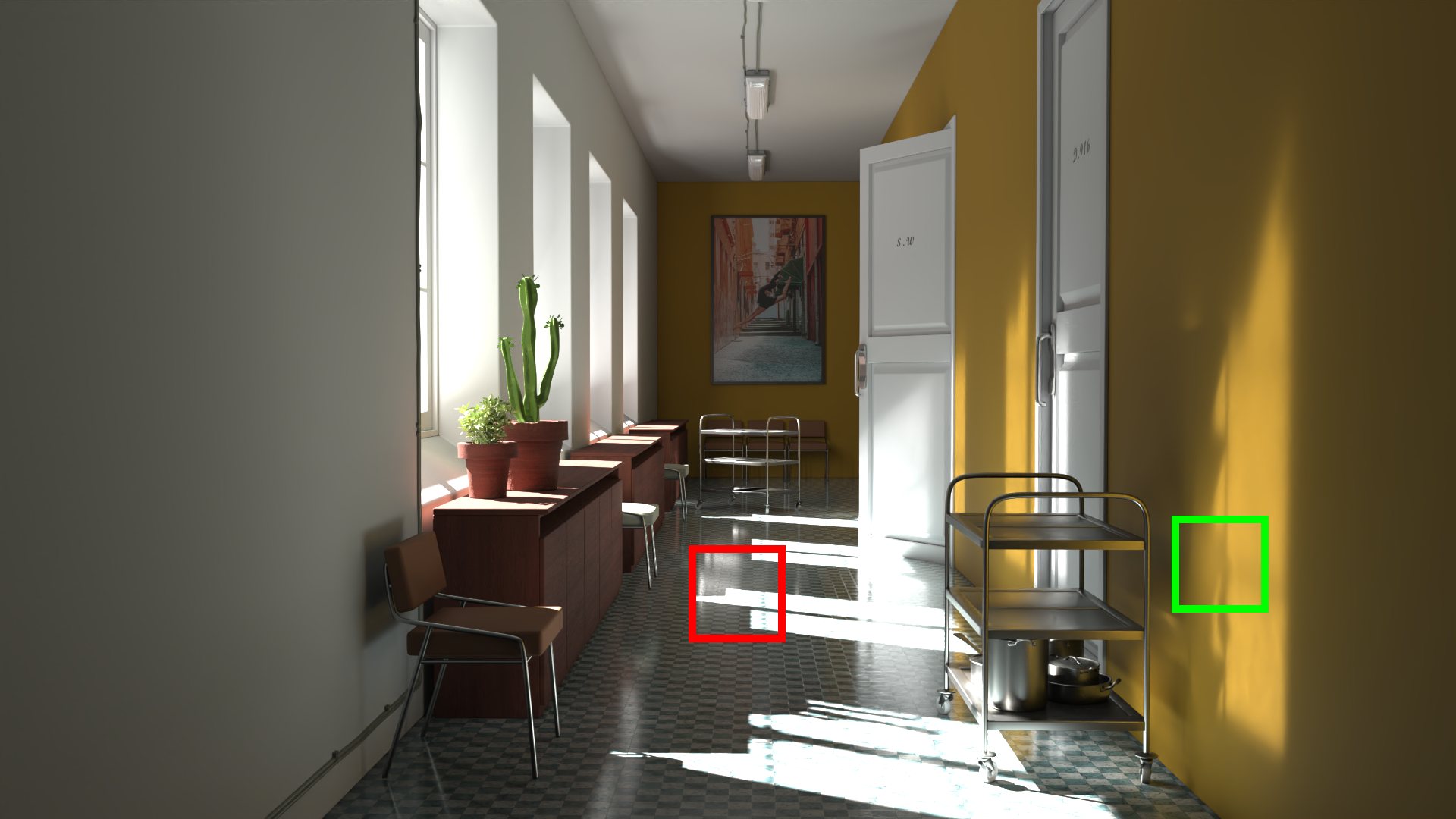}
			\end{minipage}
			&
			\begin{minipage}{0.08\linewidth}
				\includegraphics[height=1.5cm]{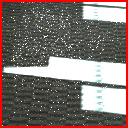}
				\newline
				\includegraphics[height=1.5cm]{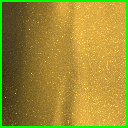}
			\end{minipage}
                &
			\begin{minipage}{0.08\linewidth}
				\includegraphics[height=1.5cm]{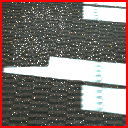}
				\newline
				\includegraphics[height=1.5cm]{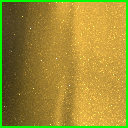}
			\end{minipage}
                &
			\begin{minipage}{0.08\linewidth}
				\includegraphics[height=1.5cm]{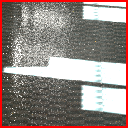}
				\newline
				\includegraphics[height=1.5cm]{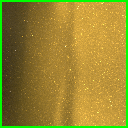}
			\end{minipage}
                &
			\begin{minipage}{0.08\linewidth}
				\includegraphics[height=1.5cm]{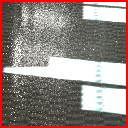}
				\newline
				\includegraphics[height=1.5cm]{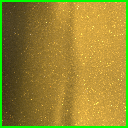}
			\end{minipage}
                &
			\begin{minipage}{0.08\linewidth}
				\includegraphics[height=1.5cm]{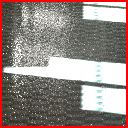}
				\newline
				\includegraphics[height=1.5cm]{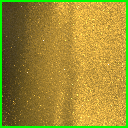}
			\end{minipage}
                &
			\begin{minipage}{0.08\linewidth}
				\includegraphics[height=1.5cm]{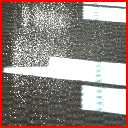}
				\newline
				\includegraphics[height=1.5cm]{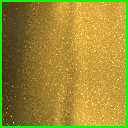}
			\end{minipage}
                &
			\begin{minipage}{0.08\linewidth}
				\includegraphics[height=1.5cm]{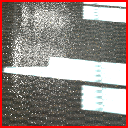}
				\newline
				\includegraphics[height=1.5cm]{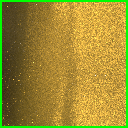}
			\end{minipage}
                \\
                &MAPE:&0.148 & 0.142 & \textbf{0.114} & 0.117 & 0.150 & 0.132 & 0.146\\
			\rotatebox[origin=c]{90}{(b) \textsc{Glossy Tube}}&
			\begin{minipage}{0.3\linewidth}
                \centering
				\includegraphics[height=3.0cm]{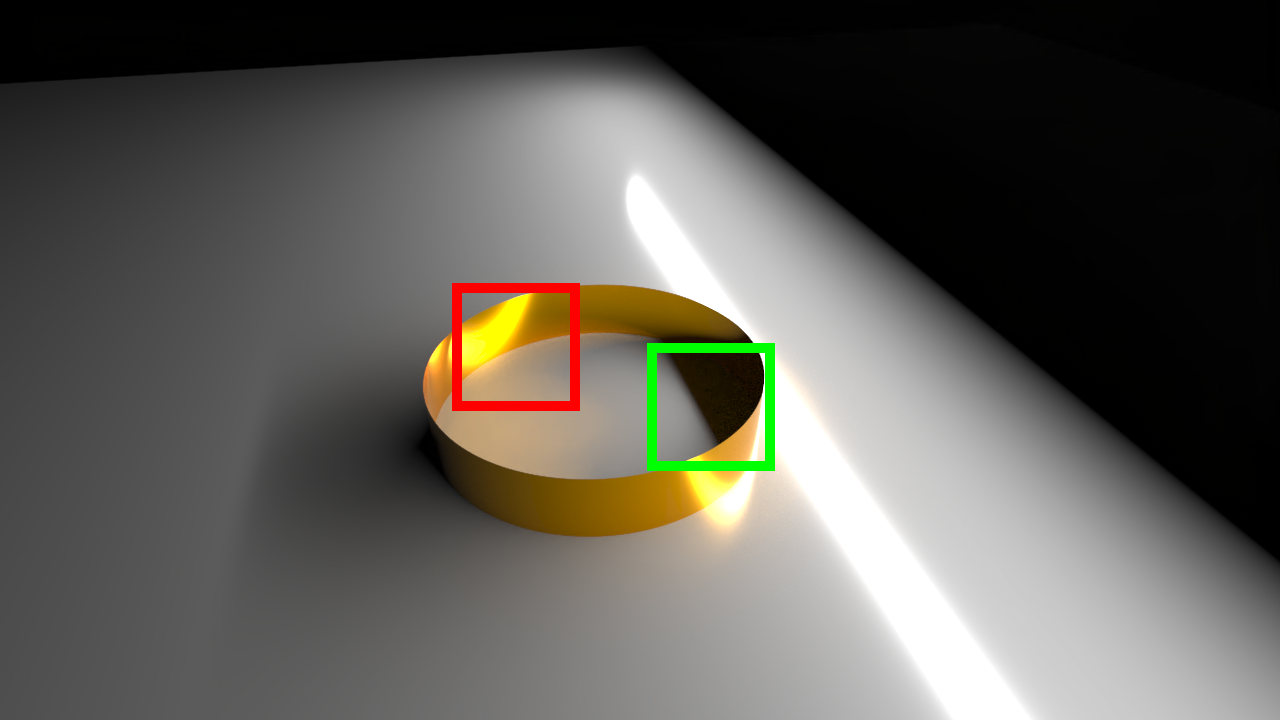}
			\end{minipage}
			&
			\begin{minipage}{0.08\linewidth}
				\includegraphics[height=1.5cm]{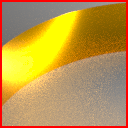}
				\newline
				\includegraphics[height=1.5cm]{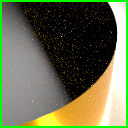}
			\end{minipage}
                &
			\begin{minipage}{0.08\linewidth}
				\includegraphics[height=1.5cm]{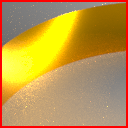}
				\newline
				\includegraphics[height=1.5cm]{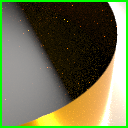}
			\end{minipage}
                &
			\begin{minipage}{0.08\linewidth}
				\includegraphics[height=1.5cm]{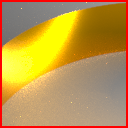}
				\newline
				\includegraphics[height=1.5cm]{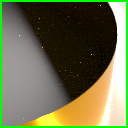}
			\end{minipage}
                &
			\begin{minipage}{0.08\linewidth}
				\includegraphics[height=1.5cm]{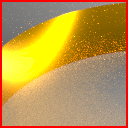}
				\newline
				\includegraphics[height=1.5cm]{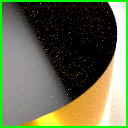}
			\end{minipage}
                &
			\begin{minipage}{0.08\linewidth}
				\includegraphics[height=1.5cm]{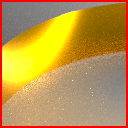}
				\newline
				\includegraphics[height=1.5cm]{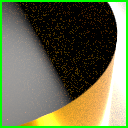}
			\end{minipage}
                &
			\begin{minipage}{0.08\linewidth}
				\includegraphics[height=1.5cm]{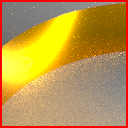}
				\newline
				\includegraphics[height=1.5cm]{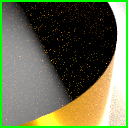}
			\end{minipage}
                &
			\begin{minipage}{0.08\linewidth}
				\includegraphics[height=1.5cm]{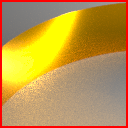}
				\newline
				\includegraphics[height=1.5cm]{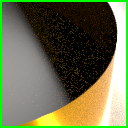}
			\end{minipage}\\
   &MAPE:&0.010 & 0.011 & \textbf{0.009} & 0.015 & 0.010 & 0.015 &0.010\\
            \end{tabular}
            \caption{A subset of rendering results in variance reduction comparison, rendered at 1024 samples per pixel (SPP). (a) Compared with incident radiance guiding, our full product guiding significantly improves results on specular surface: our ``Full Product'' result shows clear reflection of the metallic shelf, while ``Li'' or cosine weighted guiding only produces noise. MDMA and Kent distribution fail to learn an accurate distribution for narrow incident light, leading to higher variance. (b) Compared to traditional distributions such as SG, our NASG is more expressive and can learn a more accurate distribution for anisotropic lighting conditions, leading to lower variance. }
            \label{fig:representation}
        \end{figure*}
    \paragraph{Learned distribution} Our framework can easily learn a full product. In all of the test scenes, guiding with full product distribution gives lower variance. Guiding with full product helps to effectively reduce variance when rendering smooth surfaces. This is because learning full product not only improves the accuracy of the learned distribution but also helps to learn an optimal selection probability. The difference is more obvious in scenes such as \textsc{Ajar} and \textsc{Corridor}, where the floors are large surfaces with low roughness. When guiding with incident radiance distribution, the lack of BSDF distribution approximation results in many more outliers, as shown in \figref{representation} (a). 

\begin{figure*}
          \centering
  \begin{tabular}{ccccccc}
&Reference&NASG (ours)&SG&G2D&MDMA&KENT\\
    \includegraphics[height=2.0cm]{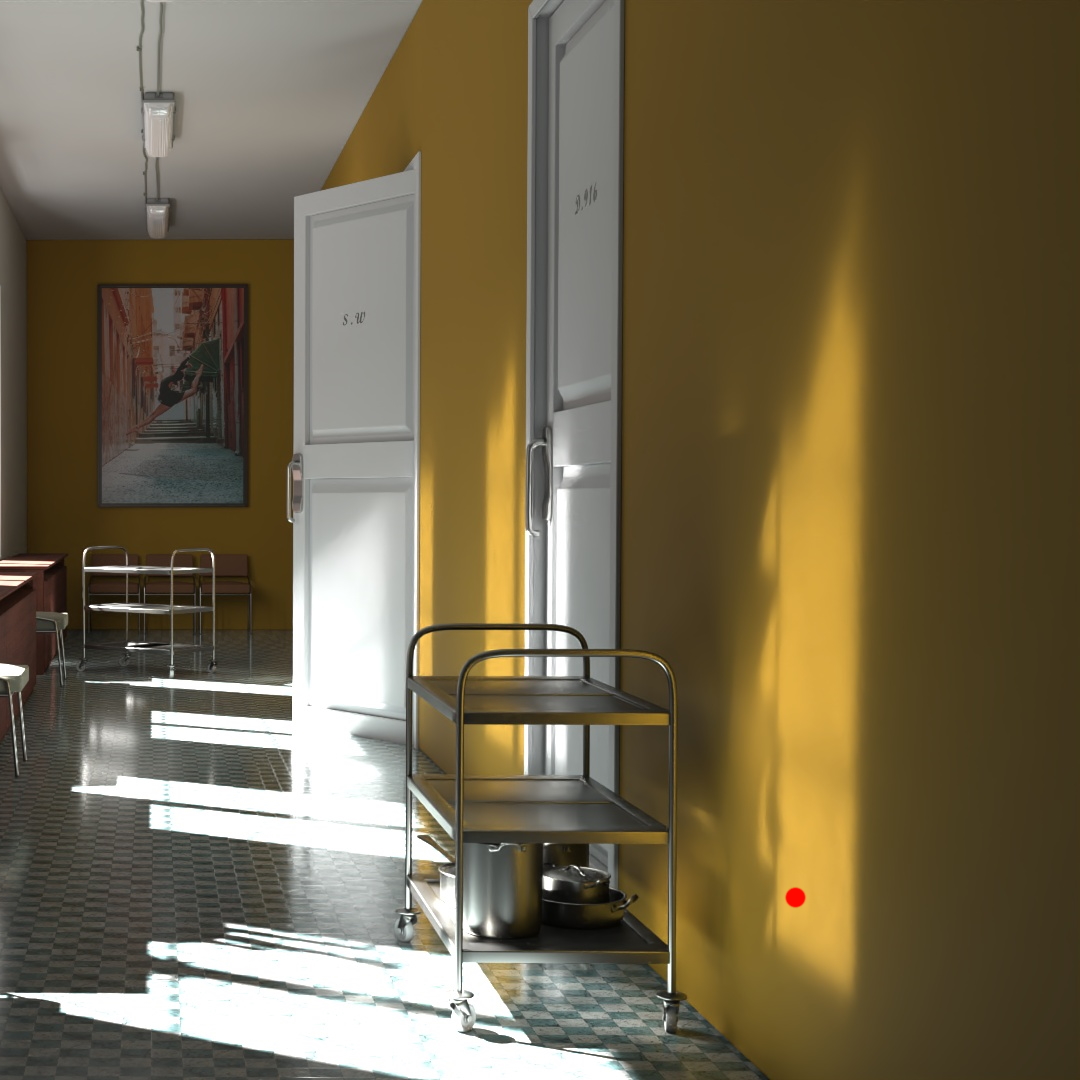}&
  \includegraphics[height=2.0cm]{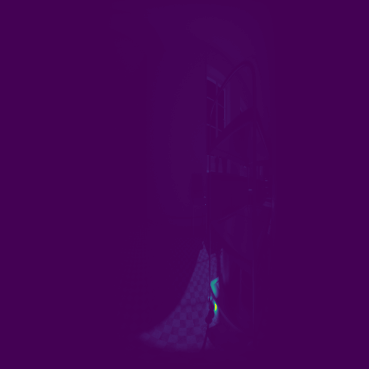}&
  \includegraphics[height=2.0cm]{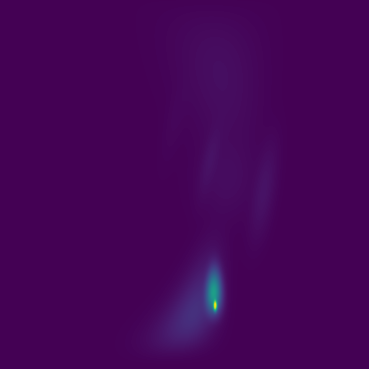}&
  \includegraphics[height=2.0cm]{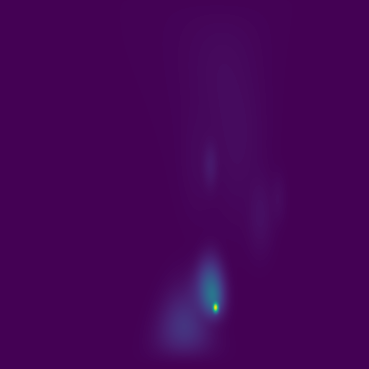}&
  \includegraphics[height=2.0cm]{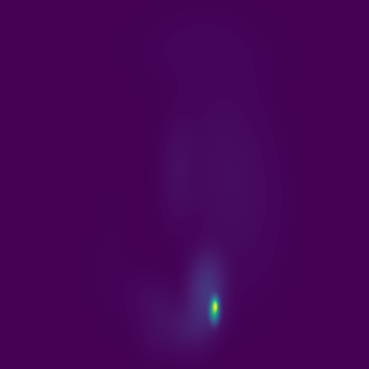}&
  \includegraphics[height=2.0cm]{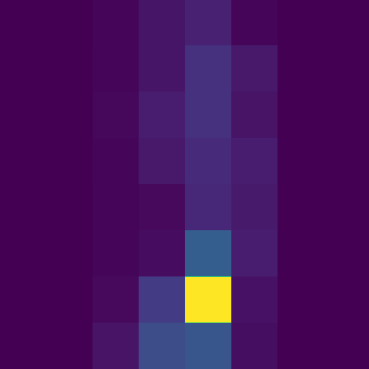}&
  \includegraphics[height=2.0cm]{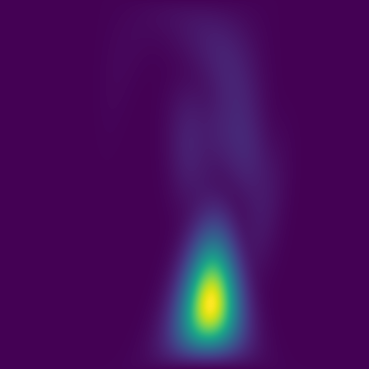}\\
  (a) \textsc{Corridor}&KL Divergence:&\bf{0.060}&0.063&0.096&0.156&0.151\\
  
  \includegraphics[height=2.0cm]{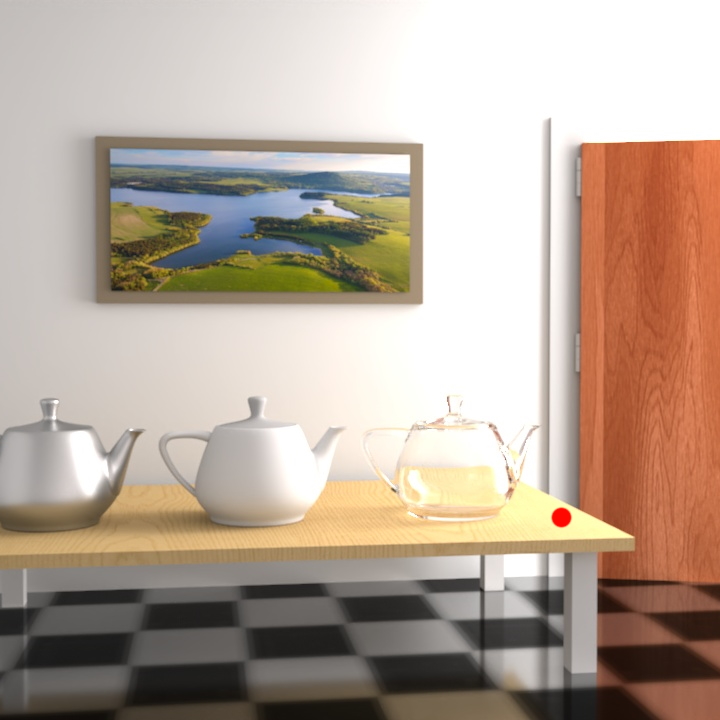}&
  \includegraphics[height=2.0cm]{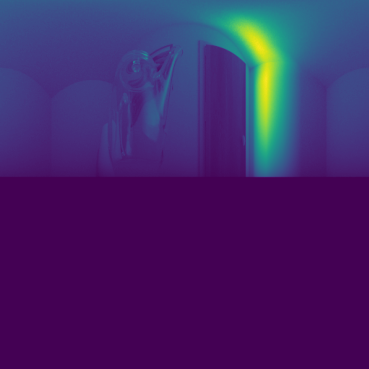}&
  \includegraphics[height=2.0cm]{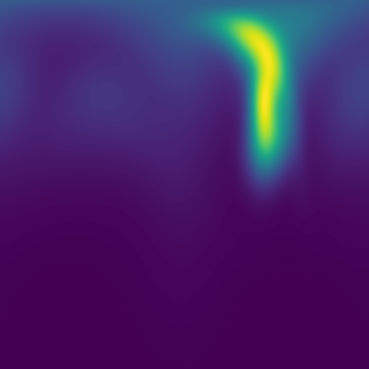}&
  \includegraphics[height=2.0cm]{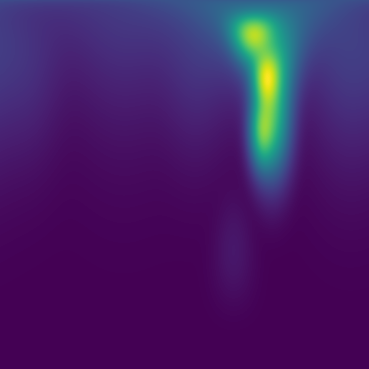}&
  \includegraphics[height=2.0cm]{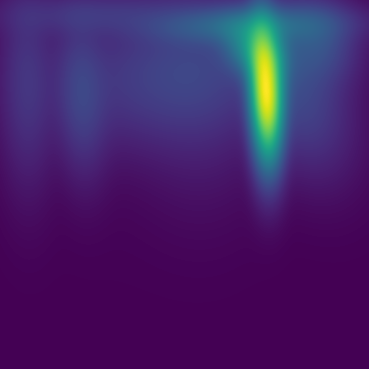}&
  \includegraphics[height=2.0cm]{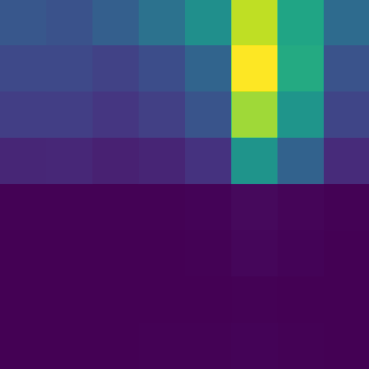}&
  \includegraphics[height=2.0cm]{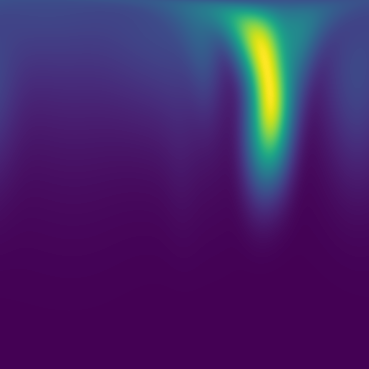}\\
  (b) \textsc{Ajar}&KL Divergence:&\bf{0.005}&0.012&0.014&0.023&0.006\\

    \includegraphics[height=2.0cm]{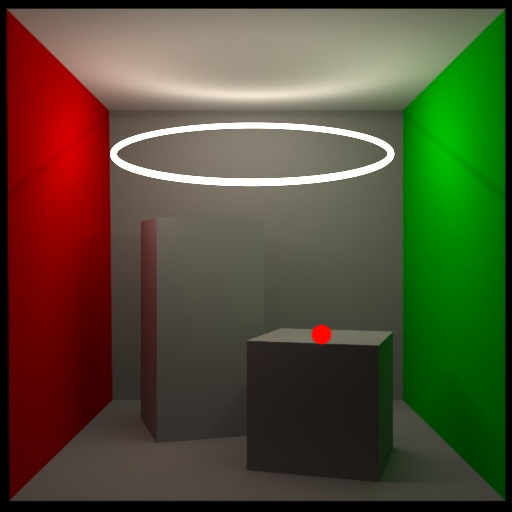}&
  \includegraphics[height=2.0cm]{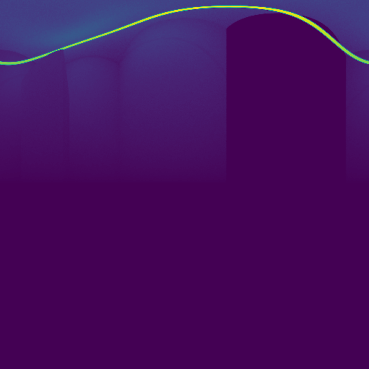}&
  \includegraphics[height=2.0cm]{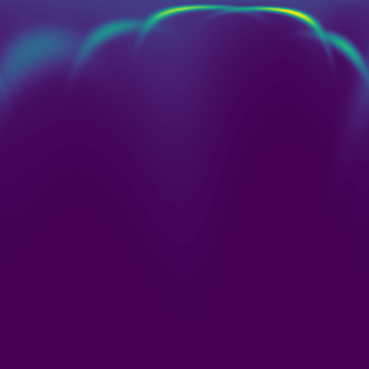}&
  \includegraphics[height=2.0cm]{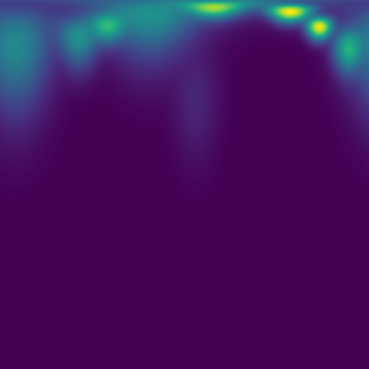}&
  \includegraphics[height=2.0cm]{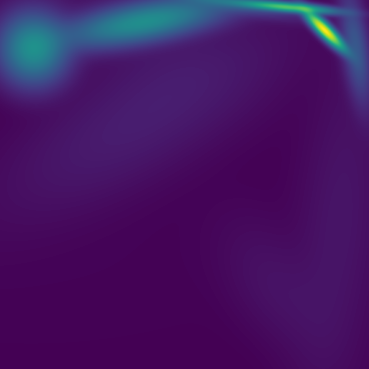}&
  \includegraphics[height=2.0cm]{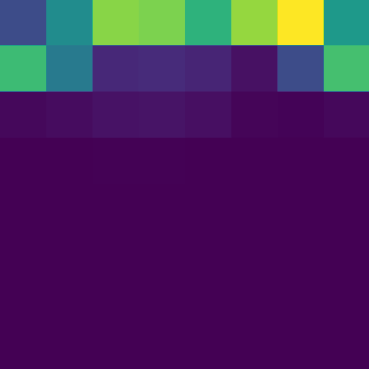}&
  \includegraphics[height=2.0cm]{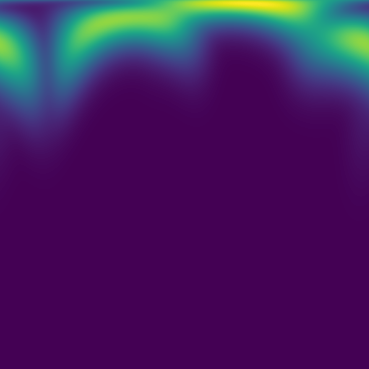}\\
  (c) \textsc{Box}&KL Divergence:&\bf{0.058}&0.078&0.098&0.112&0.087\\
  \end{tabular}
  \caption{Visualization images of learned distributions using different models, including 14-component spherical Gaussians (SG), 12-component 2D Gaussians (G2D), MDMA with 8 components for both dimensions, 8-component Kent distribution (KENT, due to the precision issue, we apply a constraint $0<\kappa<20$), and our 8-component NASG. Each row shows the learned distributions at the red-dotted locations. The reference is generated by rendering a panorama image at given locations, to which a cosine weight is applied. We measured the difference from the reference using KL-divergence (lower is better). (a) Our NASG achieves slightly more accurate distribution in an ordinary indoor indirect lighting scene. (b) \textsc{Ajar} scene features a vertical indirect lighting source, and our NASG can fit the shape well with fewer components. (c) The thin curve distribution in the \textsc{Box} scene is difficult to fit with isotropic models: the approximated curve is much thicker than the reference. In contrast, our NASG effectively preserves the thin nature of the curve, achieving lower KL-divergence. The Kent distribution fails to learn sharp distributions due the limited range of $\kappa$. However, raising the limit leads to a biased result or disrupted learning. }
  \label{fig:distribution_vis}
    \end{figure*}

    \paragraph{Distribution Models} Compared to traditional distribution models such as spherical Gaussians, our NASG model is more expressive and achieves lower variance in most scenes, and this is especially the case for scenes with anisotropic lighting conditions, such as \textsc{Ajar, Ring}, and \textsc{Glossy Tube}. In \figref{representation} (b) we show an example of a scene where the dominant indirect lighting is anisotropic. Although Kent distribution is also a spherical anisotropic distribution, we have observed two limitations: First, the computationally expensive nature of Kent distribution makes it $5\times$ slower than NASG. Second, due to the constraints imposed to mitigate the precision issue, Kent distribution can only yield a coarse approximation when the target distribution contains high-energy spots. As illustrated in \figref{representation} (a), the caustics emanating from the smooth floor are not accurately reconstructed when using Kent distribution. In addition  the rendered images, several learned distributions with different models are compared in \figref{distribution_vis}.

    \subsection{Performance}\label{sec:Performance}
    An importance sampling method is more practical when it achieves a lower level of error with the same time budget. We conducted an experiment to compare the equal-time performance of our framework with different configurations, as well as with plain path tracing, using the same set of test scenes as in our previous experiments. 
    There are several meta-parameters that can influence performance and accuracy, including the maximum sample size ($S$), number of hidden units (HU), and number of components (C). In our default configuration, we set $S = 2^{16}$, HU = 128, and C = 8. 
    We show the results under different configurations in \tabref{performance}. In general, the default configuration achieves a good balance between learning accuracy and sampling speed: 
    it achieves the lowest MAPE in four scenes out of eight. Even in the scenes where it does not perform the best,   
    its MAPE is close to the best results. 
    With 32 components (C=32), the result of \textsc{Box} has the lowest MAPE, while the sample count is only 58\% of our default configuration. 
    Actually in a simple scene such as \textsc{Box}, where performance is not a bottleneck and all methods achieve significant number of samples, a more precise distribution can be learned with more components, resulting in a lower error. However, in complex scenes where higher variance is introduced by multiple light sources or complex materials,  
    sufficient samples are required for convergence for Monte Carlo rendering. With C=32, sample count is low under limited computation budget, resulting in larger error.

        \begin{table*}[h]
    \caption{Equal-time comparison results, using a fixed 300-second budget. We compare the rendering results of plain path tracing (PT) with our framework in its default configuration, as well as several different configurations: 4 components (C=4), 16 components (C=16), 32 components (C=32), 64 hidden units per network layer (HU=64), 256 hidden units per network layer (HU=256), and a larger sample size (S=$2^{18}$).}
    \label{tab:performance}
\begin{tabular}{llrrrrrrrr}
\toprule
Scene & Metric & Default & C=4&C=16 &C=32& HU=64 & HU=256 & S=$2^{18}$ & PT \\
\midrule
\multirow{2}{*}{\textsc{Ajar}} & SPP & 1419 & 1449 & 1241 & 768 & 1590 & 1200 & 882 & 4800 \\
 & MAPE & \textbf{0.050} & 0.050 & 0.050 & 0.052 & 0.054 & 0.054 & 0.059 & 0.075 \\
\hline
\multirow{2}{*}{\textsc{Bathroom}} & SPP & 800 & 816 & 756 & 559 & 870 & 650 & 595 & 1800 \\
 & MAPE & \textbf{0.082} & 0.082 & 0.086 & 0.095 & 0.085 & 0.097 & 0.091 & 0.105 \\
\hline
\multirow{2}{*}{\textsc{Bidir}} & SPP & 2430 & 2471 & 2274 & 1372 & 2586 & 2231 & 1320 & 12300 \\
 & MAPE & 0.053 & 0.051 & 0.053 & 0.054 & 0.056 & \textbf{0.050} & 0.055 & 0.114 \\
\hline
\multirow{2}{*}{\textsc{Box}} & SPP & 3318 & 3386 & 3480 & 1912 & 3600 & 2950 & 1225 & 18500 \\
 & MAPE & 0.027 & 0.035 & 0.027 & \textbf{0.024} & 0.034 & 0.026 & 0.045 & 0.028 \\
\hline
\multirow{2}{*}{\textsc{Corridor}} & SPP & 1350 & 1377 & 1248 & 733 & 1455 & 980 & 753 & 3680 \\
 & MAPE & \textbf{0.106} & 0.110 & 0.127 & 0.122 & 0.116 & 0.122 & 0.159 & 0.175 \\
\hline
\multirow{2}{*}{\textsc{Kitchen}} & SPP & 920 & 961 & 873 & 579 & 1000 & 706 & 715 & 2380 \\
 & MAPE & 0.051 & 0.052 & 0.051 & 0.060 & \textbf{0.050} & 0.058 & 0.054 & 0.052 \\
\hline
\multirow{2}{*}{\textsc{Pool}} & SPP & 971 & 992 & 890 & 633 & 1100 & 880 & 730 & 1960 \\
 & MAPE & \textbf{0.042} & 0.043 & 0.055 & 0.052 & 0.056 & 0.066 & 0.058 & 0.305 \\
\hline
\multirow{2}{*}{\textsc{Glossy Tube}} & SPP & 2120 & 2141 & 2035 & 1431 & 2643 & 1957 & 1605 & 5215 \\
 & MAPE & 0.008 & 0.009 & 0.008 & 0.010 & \textbf{0.008} & 0.009 & 0.009 & 0.010 \\
\hline
\bottomrule
\end{tabular}
    \end{table*}

    \paragraph{Neural network overhead} In the experiment we measured the time cost of three parts of computation: forward execution for every shading point, training of the neural network, and path tracing. While changing through scenes, we found that typically 15\% of the time is used for forward execution, 35\% for training, and 50\% for path tracing.
    The training overhead is almost constant among different image resolutions, which makes our framework relatively costly when rendering small images. For example, in the \textsc{Bidir} scene, which uses a small $512\times512$ resolution, the ratio of neural network overhead appears to be larger. From our experience, increasing the training steps (i.e., more training batches in each iteration) leads to higher accuracy in the learned distribution, but with significant overhead.  We conjecture that using meta-learning to accelerate the training process would be a promising direction, and leave this as a future work.
    \paragraph{Network scale}
    Intuitively, a larger network, with either more weight parameters or more components, could achieve higher accuracy. However, as neural network computation is the primary overhead in our framework and complex scenes require a significant sample count to converge, it is still worthwhile to trade off accuracy for computation overhead given our limited time budget.
    
    \subsection{State-of-the-Art Comparison}\label{sec:sota}
\begin{figure}[h]
        \begin{tabular}{ccc}
            \includegraphics[width=0.3\linewidth]{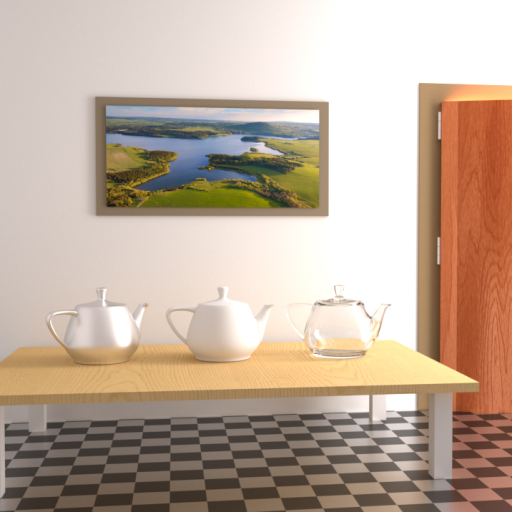} &\includegraphics[width=0.3\linewidth]{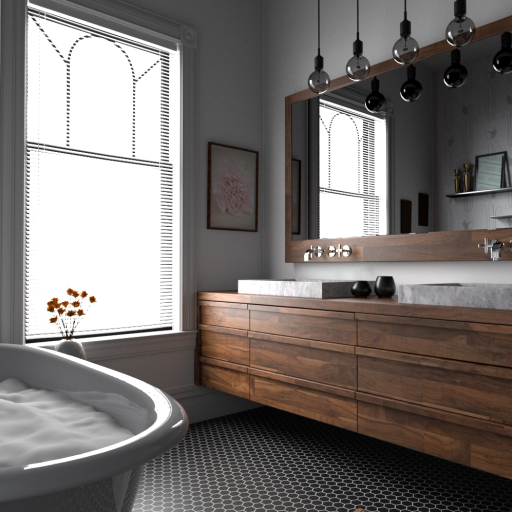}
            &\includegraphics[width=0.3\linewidth]{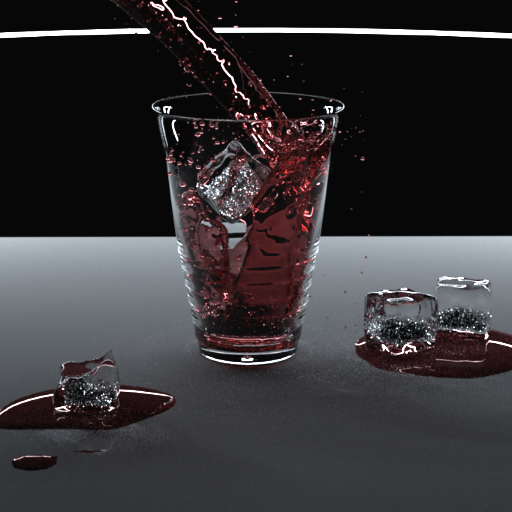}  \\
             \includegraphics[width=0.3\linewidth]{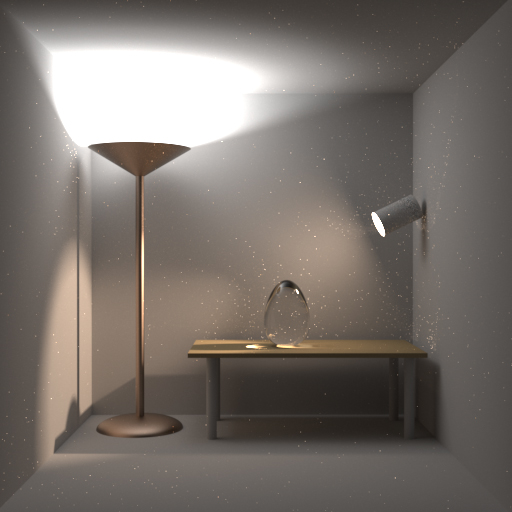} &\includegraphics[width=0.3\linewidth]{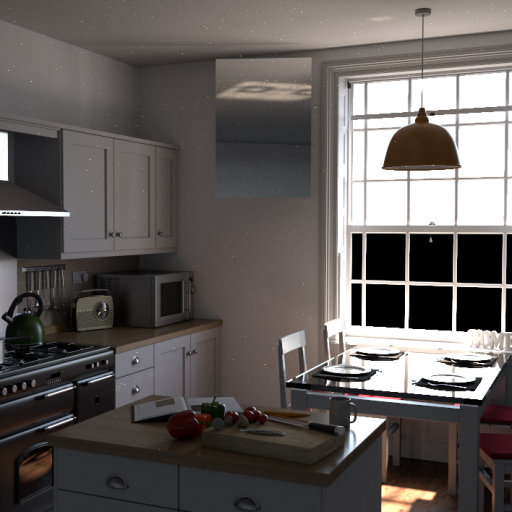} &\includegraphics[width=0.3\linewidth]{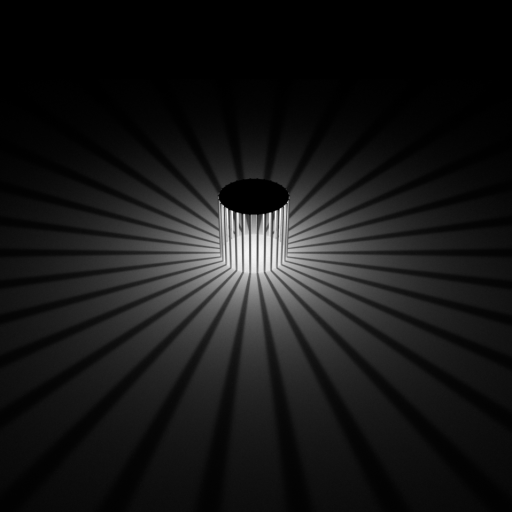} \\ \includegraphics[width=0.3\linewidth]{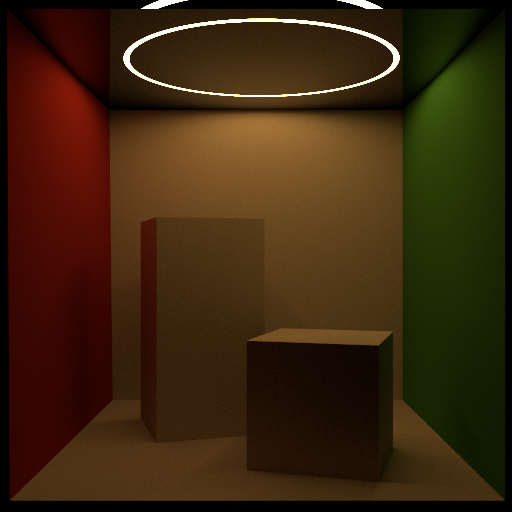} &\includegraphics[width=0.3\linewidth]{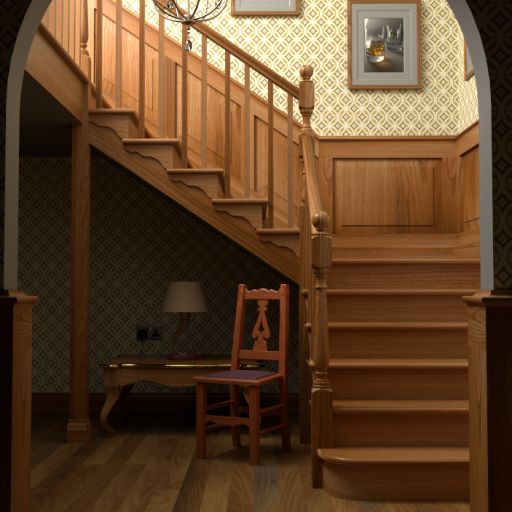} &\includegraphics[width=0.3\linewidth]{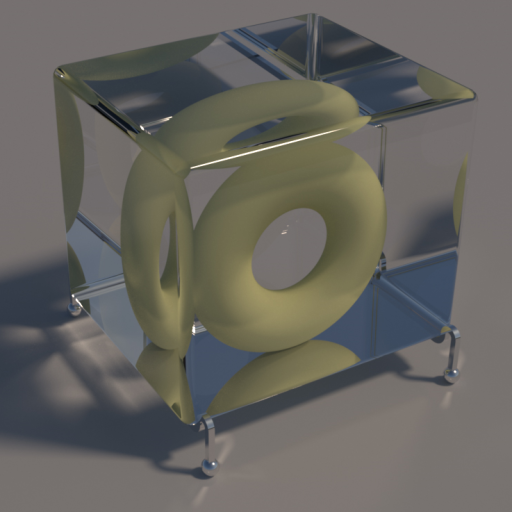}\\ 
        \end{tabular}
        \caption{Nine scenes used to evaluate proposed Mitsuba implementation with state-of-the-art techniques. From left to right: \textsc{Ajar, Bathroom, Wine, Bidir, Kitchen, Maze, Box, Staircase, Torus}. Several scenes apply different lighting setups for more indirect and anisotropic illumination.}
        \label{fig:preview_cpu}
    \end{figure}

    To compare our framework with various state-of-the-art path guiding techniques, we implemented it on the Mitsuba renderer. In our Mitsuba implementation, we set $S=2^{18}$ and $\nu = 4$ to better show the potential of our framework. The test scenes are shown in \figref{preview_cpu}, and we rendered all of the images in this experiment with two Intel Xeon E5-2680v2 20-thread CPUs (40 threads in total). We compared the results of our framework with the authors' implementations of PPG \cite{PPG19} and PAVMM \cite{PAPG}, along with our implementation of NIS \cite{MUL19}. For all techniques, we enabled NEE. 
    To render with PPG, we disabled inverse-variance-based weighting and selection probability optimizaton, while enabling tree splatting using default configuration. For PAVMM, we used BSDF product sampling when available and cosine weight sampling otherwise (the authors' implementation requires specific scene structure, which conflicts with scene setups), and we set \textit{maxSamplesPerLeafNode} to 16k, with all other options left as default. We implemented NIS following the original paper; based on our framework, we replaced the sampling-density evaluation component with 2-layer quadratic linear normalizing flow, and we added another MLP to learn selection probability. We used the encoded input of our NASG framework as extra features for the transform layer, and the feature dimensions are encoded with one-blob encoding (we used 64 bins). The training sample size and training step's configuration are aligned with our framework. 
    The major difference of our implementation from that in the original paper is that we only use lightweight MLPs, at the same scale used for our framework, rather than ResNets. The source code of our framework and our NIS implementation are given in the supplemental material. 
    
    The rendering speed of of a CPU-based neural path guiding implementation is strongly affected by the actual implementation, so we mainly compared equal-sample results rather than raw performance. 
    \subsubsection{Variance reduction of guiding method}
    To objectively evaluate the efficiency of different guiding algorithms, we first disabled orthogonal features and rendered scenes with our framework and other methods. To ensure a fair comparison, we used the same sampling parameters in all of the methods. The selection probability is set to the fixed fraction 0.5 for all methods. All of the methods have a 128-sample training budget and 384 samples for rendering (512 samples in total). We achieved this by clearing the rendering accumulation after the 128th sample, and we accumulated the results for the rest of the samples.
    
\begin{figure*}[h]
			\rotatebox{90}{\hspace{1cm}(a) MAPE result}\includegraphics[width=0.92\linewidth]{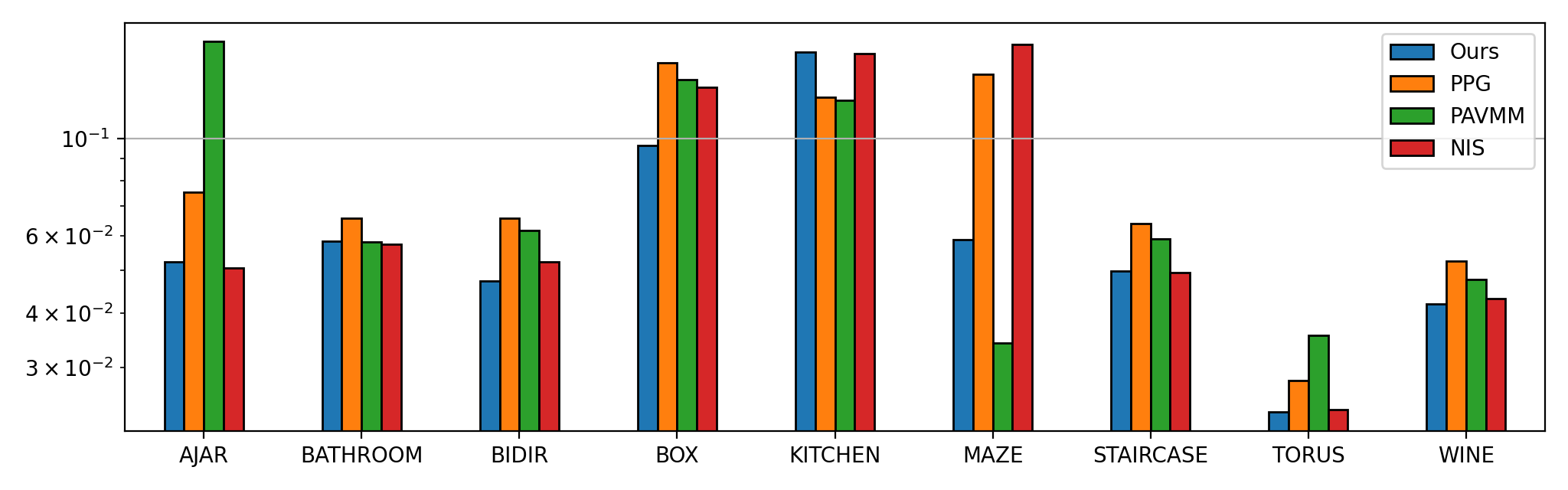}\\
		\begin{tabular}{crcccccc}
			&&PT&NASG (Ours)&PPG&NIS&PAVMM&Reference
			\\
			\rotatebox[origin=c]{90}{(b) \textsc{Box}}&
			\begin{minipage}{0.3\linewidth}
                \centering
				\includegraphics[height=3.0cm]{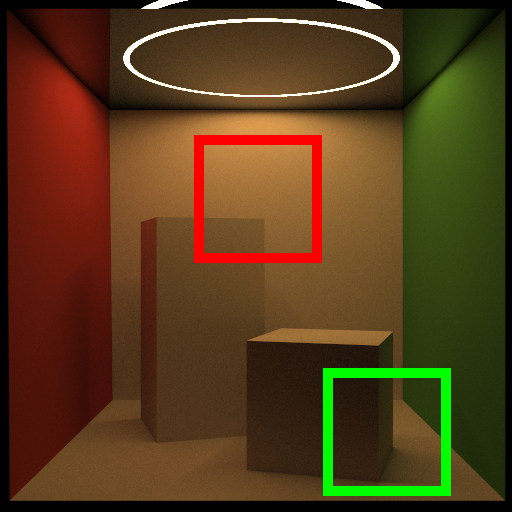}
			\end{minipage}
			&
			\begin{minipage}{0.08\linewidth}
				\includegraphics[height=1.5cm]{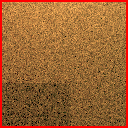}
				\newline
				\includegraphics[height=1.5cm]{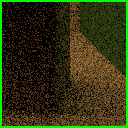}
			\end{minipage}
                &
			\begin{minipage}{0.08\linewidth}
				\includegraphics[height=1.5cm]{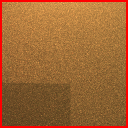}
				\newline
				\includegraphics[height=1.5cm]{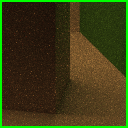}
			\end{minipage}
                &
			\begin{minipage}{0.08\linewidth}
				\includegraphics[height=1.5cm]{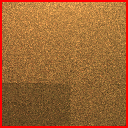}
				\newline
				\includegraphics[height=1.5cm]{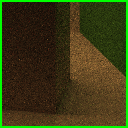}
			\end{minipage}
                &
			\begin{minipage}{0.08\linewidth}
				\includegraphics[height=1.5cm]{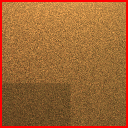}
				\newline
				\includegraphics[height=1.5cm]{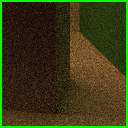}
			\end{minipage}
                &
			\begin{minipage}{0.08\linewidth}
				\includegraphics[height=1.5cm]{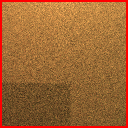}
				\newline
				\includegraphics[height=1.5cm]{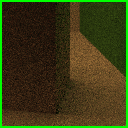}
			\end{minipage}
                &
			\begin{minipage}{0.08\linewidth}
				\includegraphics[height=1.5cm]{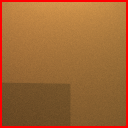}
				\newline
				\includegraphics[height=1.5cm]{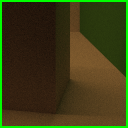}
			\end{minipage}
                \\
                \rotatebox[origin=c]{90}{(c) \textsc{Maze}}&
			\begin{minipage}{0.3\linewidth}
                \centering
				\includegraphics[height=3.0cm]{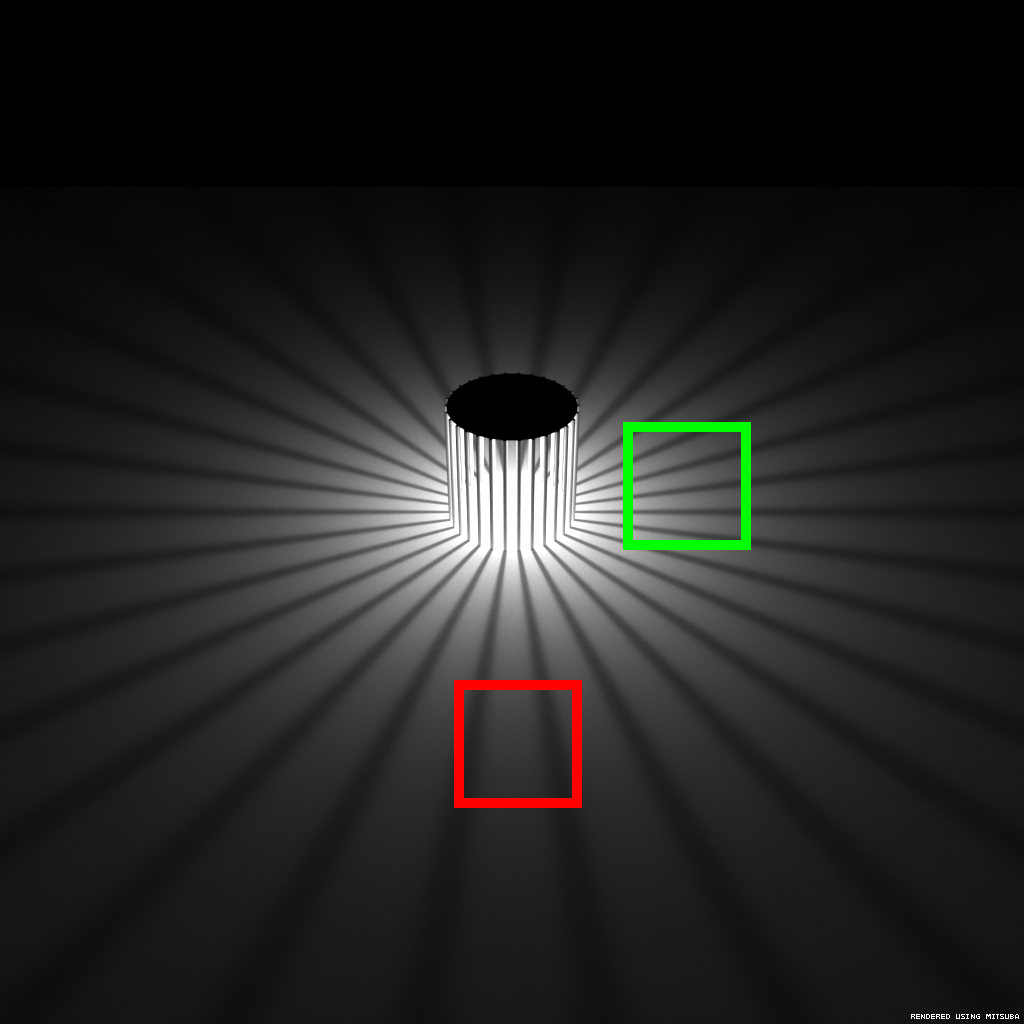}
			\end{minipage}
			&
			\begin{minipage}{0.08\linewidth}
				\includegraphics[height=1.5cm]{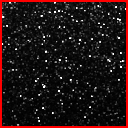}
				\newline
				\includegraphics[height=1.5cm]{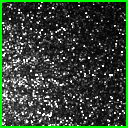}
			\end{minipage}
                &
			\begin{minipage}{0.08\linewidth}
				\includegraphics[height=1.5cm]{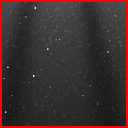}
				\newline
				\includegraphics[height=1.5cm]{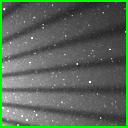}
			\end{minipage}
                &
			\begin{minipage}{0.08\linewidth}
				\includegraphics[height=1.5cm]{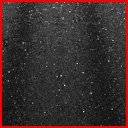}
				\newline
				\includegraphics[height=1.5cm]{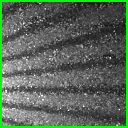}
			\end{minipage}
                &
			\begin{minipage}{0.08\linewidth}
				\includegraphics[height=1.5cm]{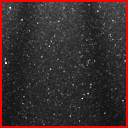}
				\newline
				\includegraphics[height=1.5cm]{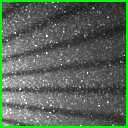}
			\end{minipage}
                &
			\begin{minipage}{0.08\linewidth}
				\includegraphics[height=1.5cm]{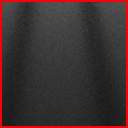}
				\newline
				\includegraphics[height=1.5cm]{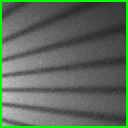}
			\end{minipage}
                &
			\begin{minipage}{0.08\linewidth}
				\includegraphics[height=1.5cm]{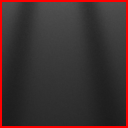}
				\newline
				\includegraphics[height=1.5cm]{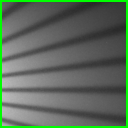}
			\end{minipage}
                \\
                \rotatebox[origin=c]{90}{(d) \textsc{Ajar}}&
			\begin{minipage}{0.3\linewidth}
                \centering
				\includegraphics[height=3.0cm]{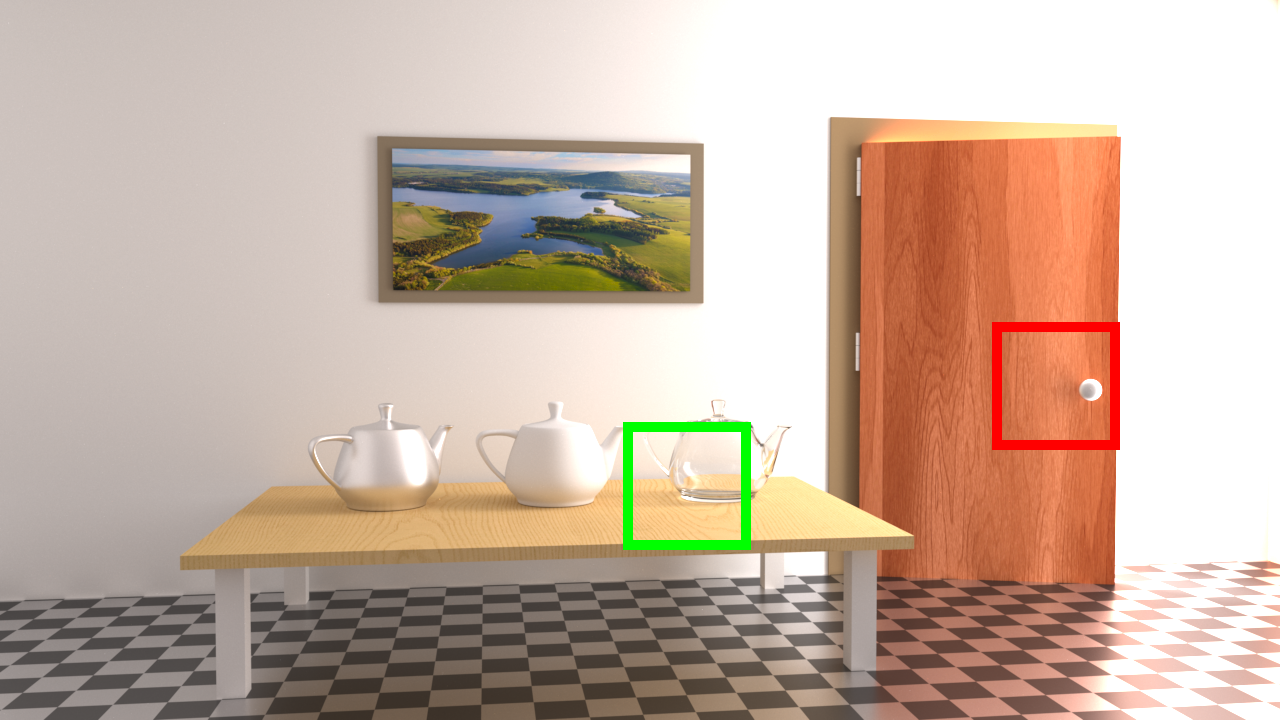}
			\end{minipage}
			&
			\begin{minipage}{0.08\linewidth}
				\includegraphics[height=1.5cm]{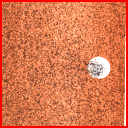}
				\newline
				\includegraphics[height=1.5cm]{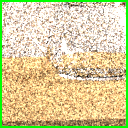}
			\end{minipage}
                &
			\begin{minipage}{0.08\linewidth}
				\includegraphics[height=1.5cm]{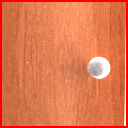}
				\newline
				\includegraphics[height=1.5cm]{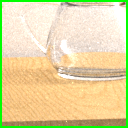}
			\end{minipage}
                &
			\begin{minipage}{0.08\linewidth}
				\includegraphics[height=1.5cm]{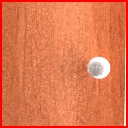}
				\newline
				\includegraphics[height=1.5cm]{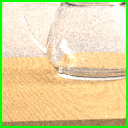}
			\end{minipage}
                &
			\begin{minipage}{0.08\linewidth}
				\includegraphics[height=1.5cm]{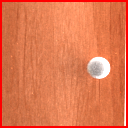}
				\newline
				\includegraphics[height=1.5cm]{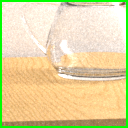}
			\end{minipage}
                &
			\begin{minipage}{0.08\linewidth}
				\includegraphics[height=1.5cm]{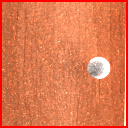}
				\newline
				\includegraphics[height=1.5cm]{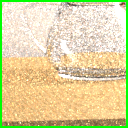}
			\end{minipage}
                &
			\begin{minipage}{0.08\linewidth}
				\includegraphics[height=1.5cm]{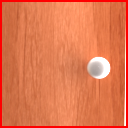}
				\newline
				\includegraphics[height=1.5cm]{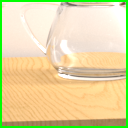}
			\end{minipage}
                \\
            \end{tabular}
		\caption{Results of state-of-the-art comparison, visualized in logarithmic scale (lower is better). We render a variety of 3D scenes with 512 SPP, guided with different techniques including our framework, PPG \cite{PPG19}, PAVMM \cite{PAPG}, and NIS \cite{MUL19}. Each technique uses the authors' implementations without modification except NIS. A subset of rendering results that demonstrates: (b) The \textsc{Box} scene, which features a circle luminary lighting the scene via a glass ceiling. The lighting is indirect and anisotropic, which is challenging for isotropic models such as vMF used in PAVMM. Our NASG helps to fit the distribution accurately and achieve lower variance. (c) The \textsc{Maze} scene with strong indirect lighting from a cage, which is challenging for PPG due to the parallax issue, while PAVMM achieves a clean result with its parallax-aware fitting algorithm, closely followed by ours. However, in (d) the \textsc{Ajar} scene, PAVMM has difficulty learning and fitting accurate distribution due to anisotropic indirect lighting and the near-specular floor, while our framework guides robustly. Although every method has its advantages in special circumstances, our framework performs well in general, making it an attractive option as a path-guiding technique.}
		\label{fig:guiding_compare}
	\end{figure*}

\begin{figure*}
        \begin{tabular}{ccc}
             \includegraphics[height=5cm]{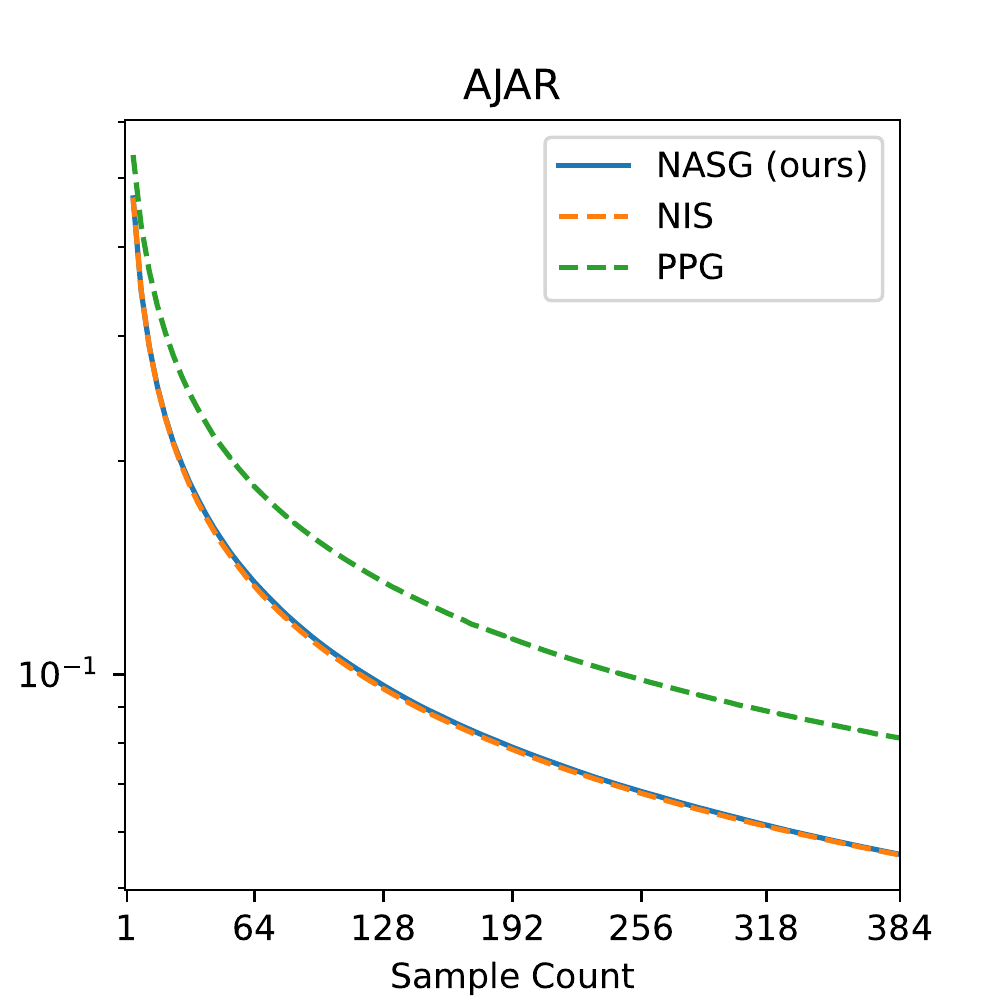}&
             \includegraphics[height=5cm]{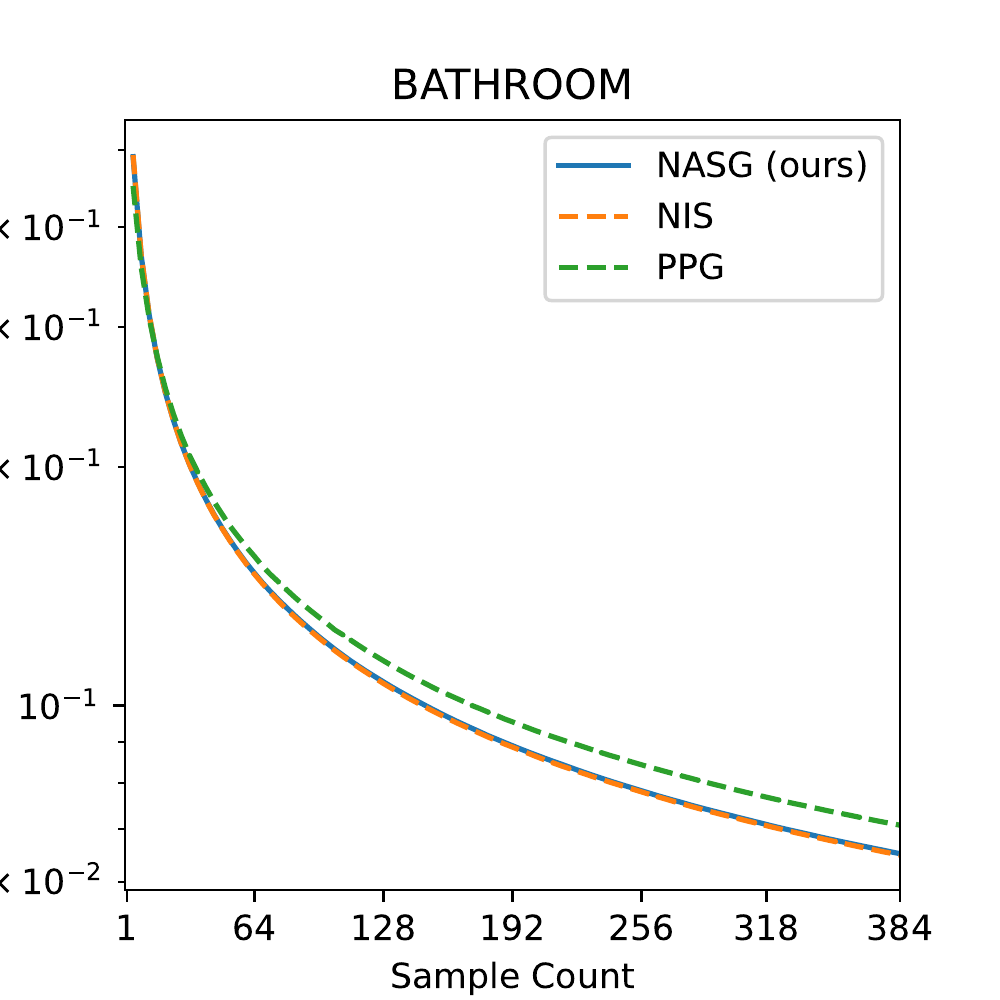}&
             \includegraphics[height=5cm]{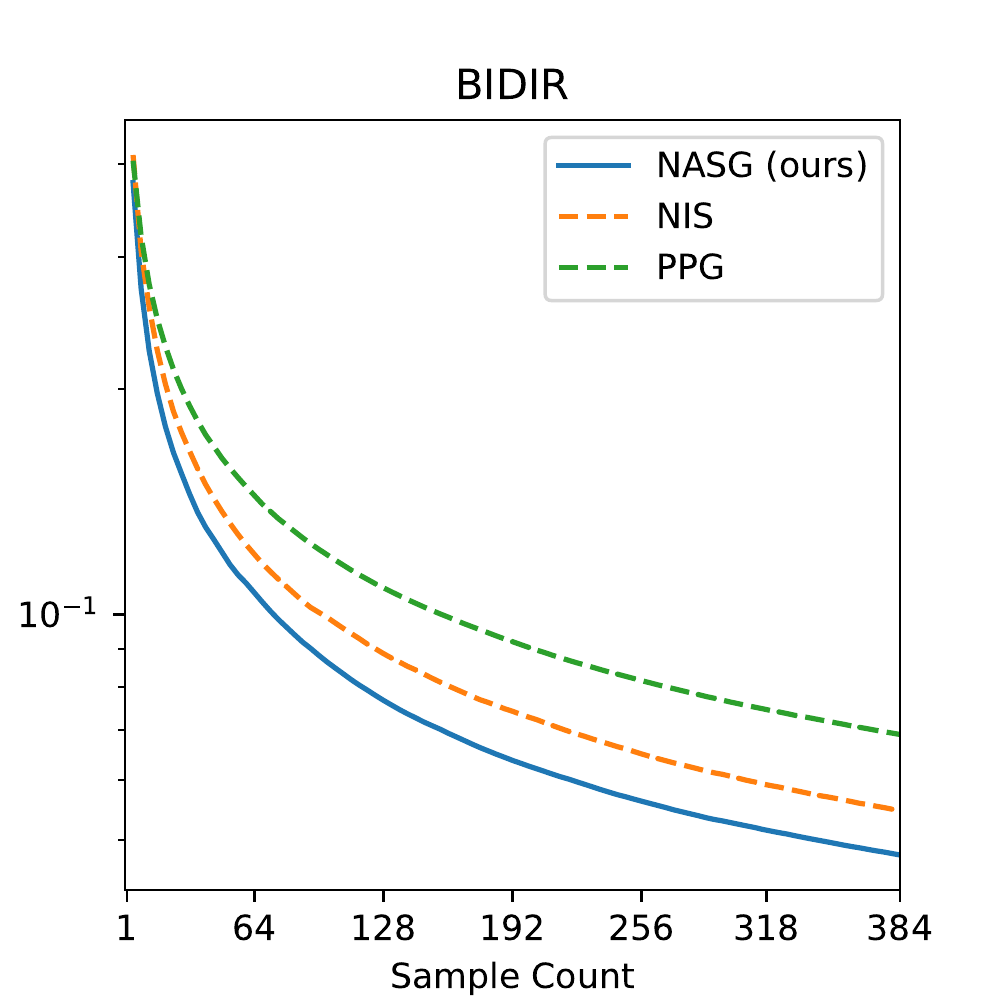}  
             \\
             \includegraphics[height=5cm]{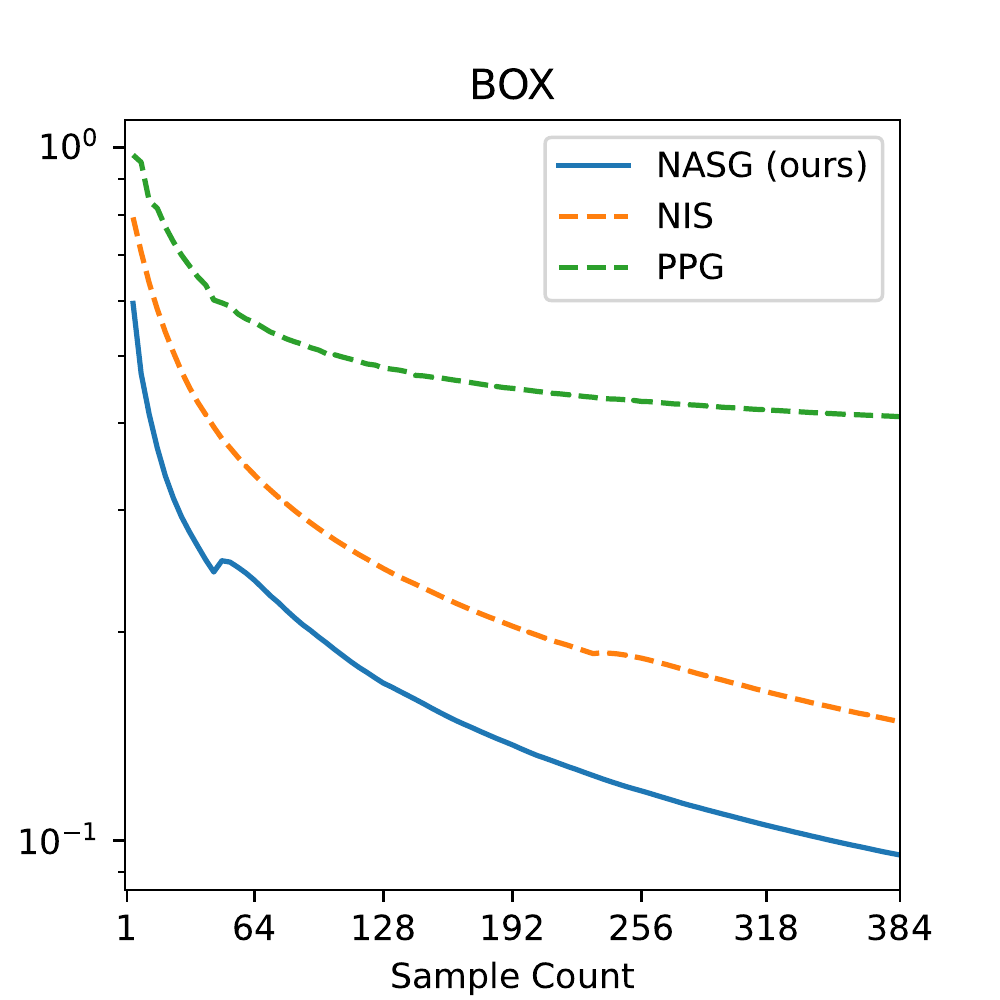}&
             \includegraphics[height=5cm]{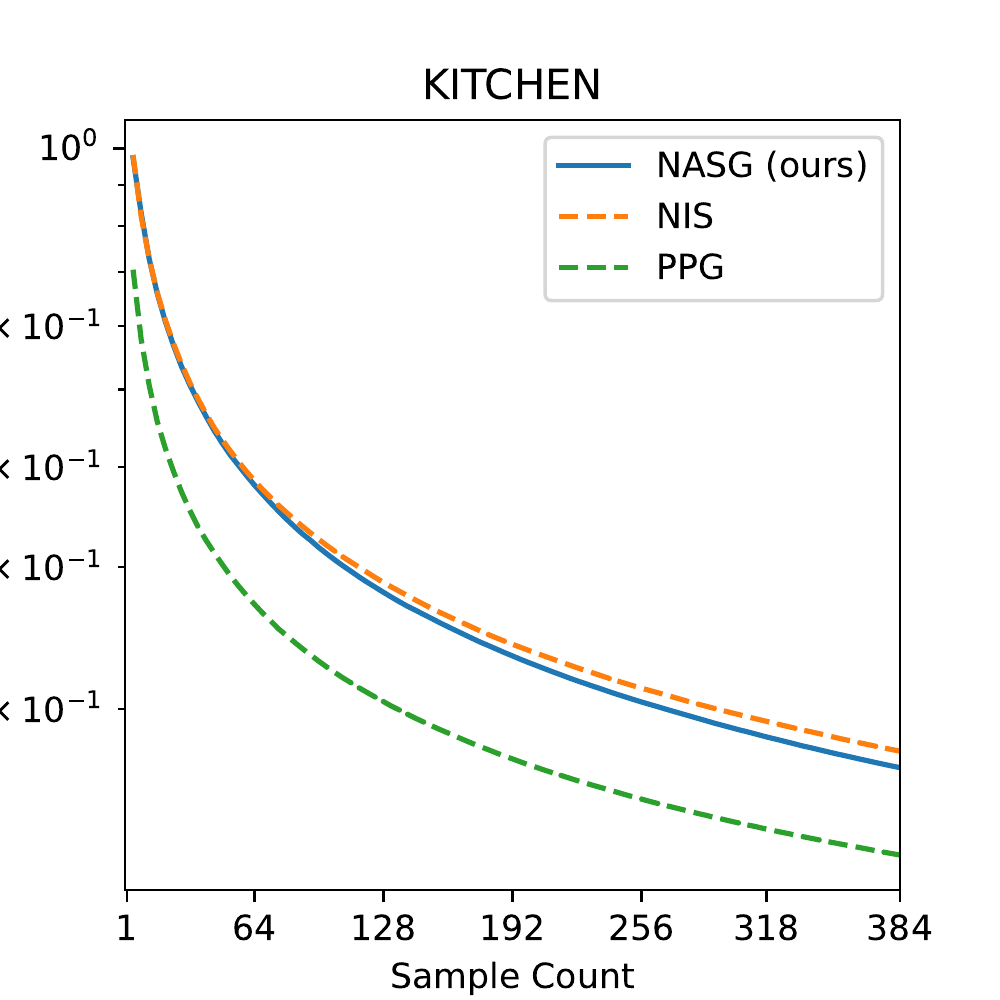}&
             \includegraphics[height=5cm]{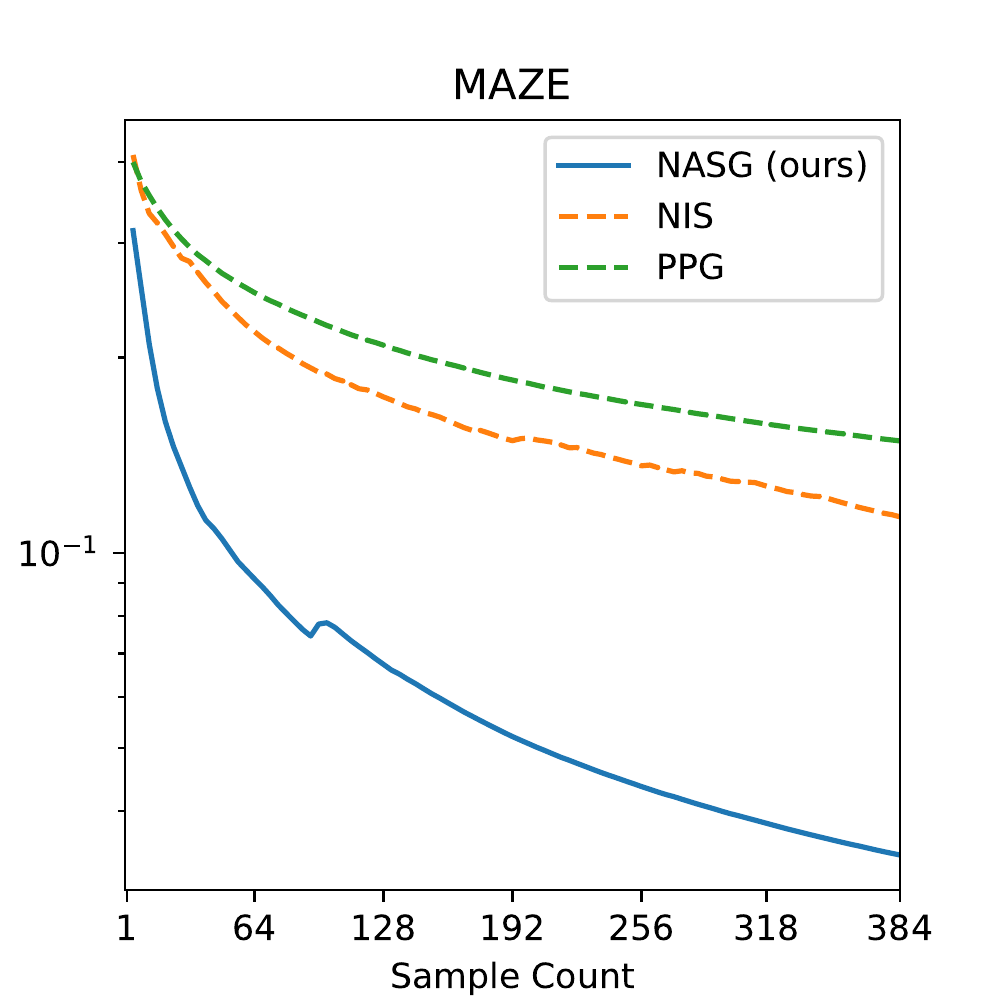}  
             \\
             \includegraphics[height=5cm]{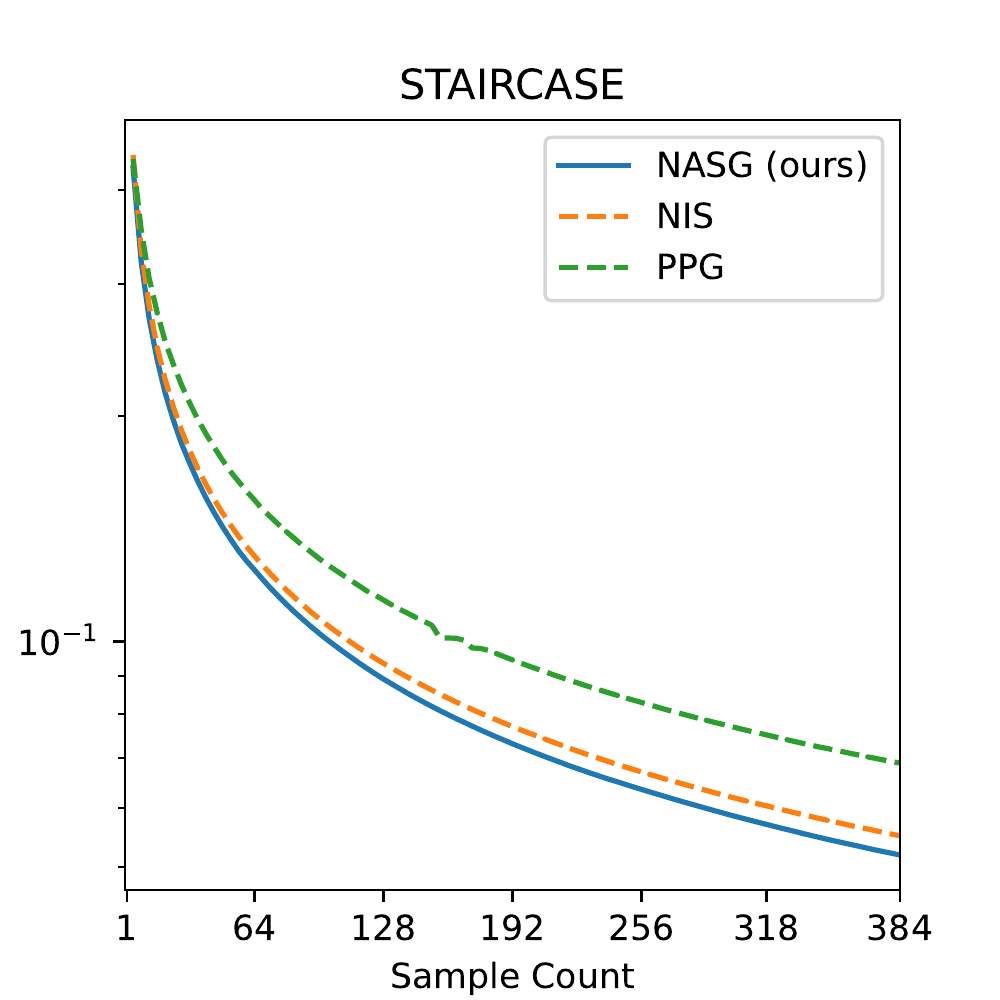}&
             \includegraphics[height=5cm]{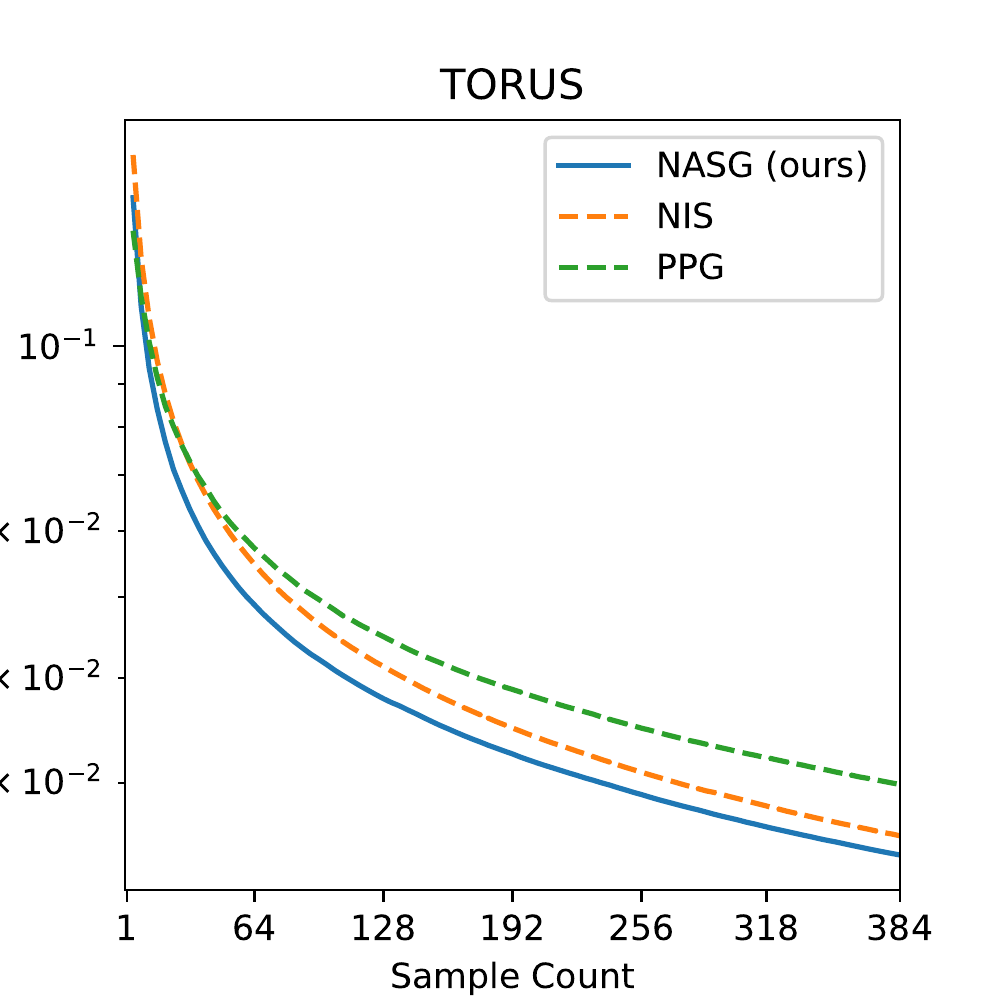}&
             \includegraphics[height=5cm]{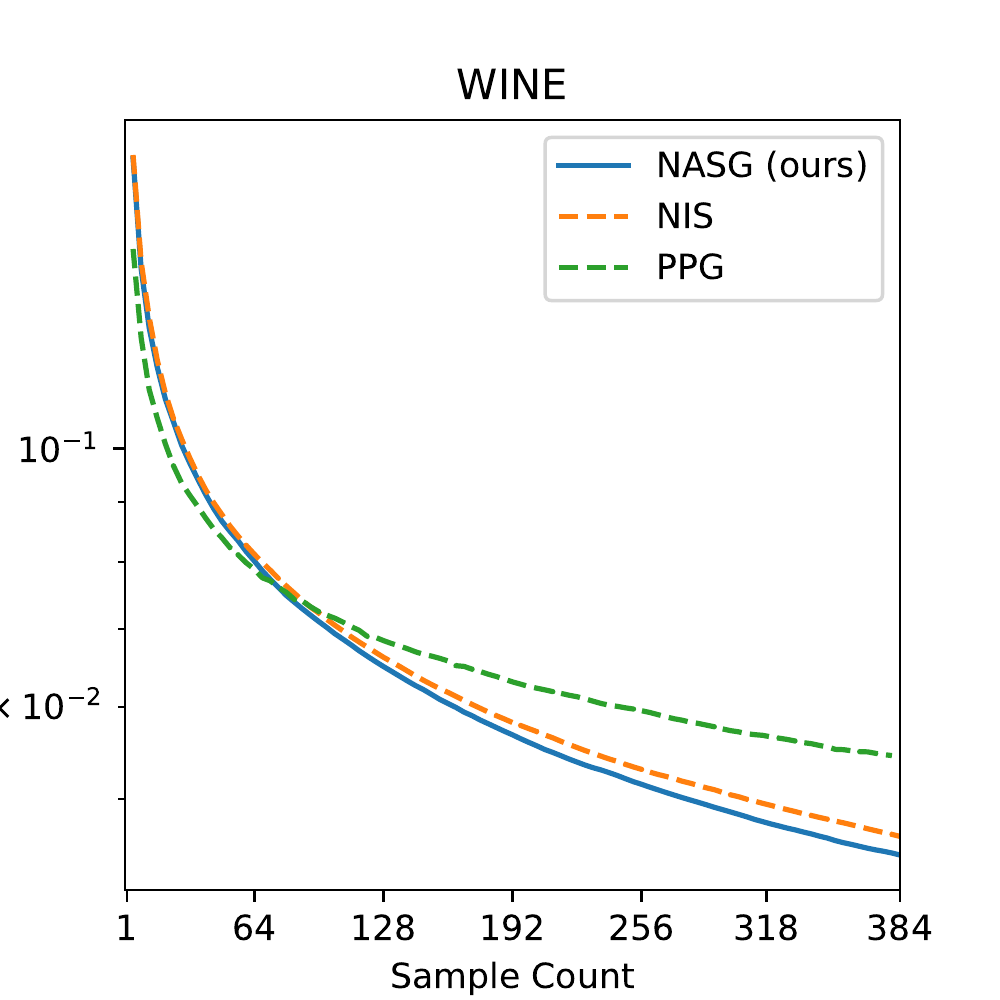} 
        \end{tabular}
        \caption{Convergence graphs of all experiment scenes. PAVMM result is not reported due to a different mechanism of rendering. All of the techniques show similar convergence behavior due to the characteristics of importance sampling, while in general, our framework learns a more accurate distribution to achieve faster convergence. In \textsc{Box} and \textsc{Maze}, we observe jitter on NASG's convergence. This can be explained as an attempt to get out of the local minima by the optimizer, with the relatively large learning rate. In the \textsc{Wine} experiment, neural path guiding methods start from higher variance but quickly surpass PPG due to the progressive learning scheme.}
        \label{fig:convergence_graph}
    \end{figure*}

    \figref{guiding_compare} reports statistical results and \figref{convergence_graph} shows the convergence behavior during the experiment. From the results, we can observe that each previous work has both pros and cons; consequently, one technique may perform very well in one kind of scene but poor in others (details discussed below). Our framework, on the other hand, while achieving overall low error also provides a more balanced performance in all test scenes.

        \paragraph{Comparison with PPG} PPG uses quad-tree for directional representation, which can closely model distributions with multiple high-energy points. Hence, in the \textsc{Torus} scene it achieves low variance, while PAVMM struggles due to the limited components used. However, in the \textsc{Maze} scene, the parallax issue makes the PPG result very noisy, as shown in \figref{guiding_compare} (c). Parallax error occurs because PPG's spatial partition algorithm has difficulty learning from narrow and close incident radiance. Our framework learns the continuous spatial-varying distribution and does not suffer from such issues.
    \paragraph{Comparison with PAVMM} PAVMM performs generally better than PPG. With its parallax-aware fitting method, it learns distributions in the \textsc{Maze} scene most accurately. 
    However, PAVMM produces a rather noisy result in the \textsc{Ajar} scene, as shown in \figref{guiding_compare}~(d). This is due to two reasons: First, the major illumination is a line-shaped lit surface on the wall at the right side of the scene (lit by light source outside the door via a small crack), which is anisotropic and difficult for vMF to fit. Second, the fitting algorithm of PAVMM encounters a challenge in modeling the indirect lighting from the specular checkbox floor. In contrast, our proposed framework exhibits greater robustness by leveraging NASG's ability to accurately capture the anisotropic distribution with a parsimonious set of components, thus achieving low variance.
    \paragraph{Comparison with NIS} With the same configuration of training samples and steps, NIS in general achieves similar variance, except in the \textsc{Maze} scene (as shown in \figref{guiding_compare}~(c), in which the learning of the 2D distribution suffers from discontinuity at the edge of the azimuth axis). Compared to the results reported in the original NIS paper \cite{MUL19}, our NIS implementation has a slightly higher variance. We attribute this to two main factors: (a) Our implementation employs a lightweight 2-layer MLP-based transform as opposed to the original, relatively more complex, 4-layer ResNet-based version and (b) we limit our training to a subset of samples at each iteration. However, even with this relatively lightweight implementation, NIS takes roughly $2\times$ longer than our framework: in \tabref{rendering_time} we compare the rendering time of NIS and our NASG framework. The extra overhead is due to the usage of coupled networks, and the normalizing flow needs to be evaluated two times at each non-delta shading point (one for scattering sampling and the other for NEE).  
   The intrinsic characteristics of normalizing flow as an implicit density model, characterized by the utilization of multi-layer transformations, imply that a larger network architecture may be imperative to effectively capture complex distributions with accuracy. Moreover, it is plausible that a greater number of training samples or iterations might be essential to ensure convergence. These adjustments are likely requisite for attaining the results reported in the original paper. 
    However, they also inherently bring a significant increase in computational overhead. In other words, with limited computation resources, our framework can learn more accurate distributions.
    \begin{table}[h]
    \caption{Rendering time (minutes) of proposed NASG framework and NIS implementation in Mitsuba. NIS uses a normalizing flow to model distribution, which takes roughly twice as much time to calculate, compared to our explicit NASG model.}
    \label{tab:rendering_time}
        \begin{tabular}{c  c  c}
        \toprule
            Scene&NASG (ours)& NIS \\\hline
             \textsc{Ajar}&188.8m&520.4m\\
             \textsc{Bathroom}&56.2m&166.8m\\
             \textsc{Bidir}&35.6mm&89.6m\\
             \textsc{Kitchen}&50.1m&163.5m\\ 
             \textsc{Maze}&66.4m&84.42m\\
             \textsc{Box}&12.4m&42.7m\\ 
             \textsc{Staircase}&46.3m&217.2m\\ 
             \textsc{Torus}&27.3m&109.2m\\ 
             \textsc{Wine}&32.2m&128.4m\\
             \hline
             \bottomrule
        \end{tabular}
    \end{table}

    \subsubsection{System comparison}
    Our neural path guiding framework integrates two features to further reduce sampling variance, namely, selection probability optimization and progressive sample re-weighting. PPG \cite{PPG19} proposed similar features yet achieved them with different approaches. To compare the sampling efficiency of fully integrated systems, we render the above scenes with all features \textit{enabled}. For PPG \cite{PPG19} we additionally set the sample budget to 511 in order to benefit from inverse-variance-weighted blending. While PAVMM \cite{PAPG} does not implement similar features, 
    similar improvement could be expected if the features were implemented with it. 

    As shown in \tabref{system_compare}, by enabling selection probability optimization and sample re-weighting, the MAPE of the results produced by both methods decreases in a generally varying range. PPG achieves a more obvious boost, which we attribute to its lack of product guiding: selection probability optimization thus has a stronger impact on reducing variance. While both methods are further improved by these features, it is also notable that our framework can optimize a selection probability almost for free, making our framework an attractive option when integrating it into a high-performance rendering system.

\begin{table*}
\caption{MAPE of rendering results, with and without enabling selection probability optimization and sample re-weighting schemes, in PPG and proposed framework. A general improvement can be observed in most cases for both methods.}
\centering
\begin{tabular}{c|ccc|ccc}
\toprule
& \multicolumn{3}{c|}{PPG} & \multicolumn{3}{c}{NASG (ours)} \\ \cline{2-7}
& Guiding & + Features & Improvement (\%) & Guiding & + Features & Improvement (\%) \\ \hline

\textsc{Ajar} & 0.075 & 0.078 & -2.83 & 0.052 & 0.050 & 4.77 \\ 
\textsc{Bathroom} & 0.066 & 0.060 & 8.17 & 0.058 & 0.060 & -3.13 \\       
\textsc{Bidir} & 0.066 & 0.061 & 7.84 & 0.047 & 0.042 & 10.73 \\
\textsc{Box} & 0.149 & 0.109 & 26.55 & 0.096 & 0.094 & 2.51 \\
\textsc{Kitchen} & 0.124 & 0.111 & 10.88 & 0.157 & 0.150 & 4.76 \\ 
\textsc{Maze} & 0.140 & 0.077 & 45.19 & 0.059 & 0.050 & 15.58 \\
\textsc{Staircase} & 0.064 & 0.055 & 14.04 & 0.050 & 0.048 & 4.16 \\       
\textsc{Torus} & 0.028 & 0.025 & 12.19 & 0.024 & 0.024 & -0.15 \\
\textsc{Wine} & 0.052 & 0.049 & 6.43 & 0.042 & 0.040 & 5.41 \\
\hline
\bottomrule
\end{tabular}
\label{tab:system_compare}
\end{table*}

	\section{Discussion}

\subsection{Performance in GPU Path Tracing}
A GPU-based brute-force path tracer can be $20\times$ faster than a CPU-based one. If an importance sampling method provides low per-sample variance yet high overhead, it cannot defeat a brute-force path tracer on GPU. As a network-based guiding method, the major overhead is the network itself, therefore, we must adopt a sampling strategy that can be implemented in a small network. This motivates us to find a solution for learning an explicit distribution, rather than an implicit model such as one based on a normalizing-flow. Despite its powerful computation ability, a GPU's large memory bandwidth can only be leveraged by carefully designed high-concurrency algorithms. This raises the necessity of reducing code branching as much as possible, which conflicts with the nature of path tracing. Otherwise, it suffers from a much more obvious memory latency than CPU, which causes many CPU algorithms to perform relatively slowly on the GPU (\eg PPG's quad-tree approach requires multiple random accesses to GPU memory, thus causing long stalls). Recently, with hardware improvement, we see some acceleration hardware that is specifically designed for neural networks, which our framework could potentially benefit from.
\subsection{Learning rate}
In our GPU implementation, we used a relatively large learning rate (0.002). This decision was strategically aimed at enabling our model to swiftly navigate the parameter space, thus facilitating rapid convergence. This feature is particularly crucial in leveraging early samples to their fullest potential in our context. However, a large learning rate limits our model's ability to reach the deepest minima. That being said, at the latter stage of training, the resulting ``jitters'' remain close to the ground truth, ensuring that our learned distributions are valid approximations. We leave the improvement of optimization strategy as future work.
    
\subsection{Limitations and Future Work}\label{sec:Discussion}
Although in general our configuration for training works well, sometimes we encounter cases where we need to use a smaller learning rate to avoid degenerated distributions. Such degenerated distributions have been observed only a few times when rendering caustics. They could be due to the complexity of the distribution, where only a small amount of light hits the point from a narrow direction. When the early samples based on BSDF fail to produce a representative initial distribution, the system could fail to sample important directions, which could eventually lead to degeneration. Adopting an adaptive learning strategy could possibly improve robustness - we leave this as a future work.

	Another interesting extension of our framework could be path guiding in scenes with participating media. This could be achieved in theory (similar to what was done previously \cite{VPG}),  yet the ability to learn 3D distributions remains an area to be further investigated.

	\section{Conclusion}
 In this paper we proposed an effective online neural path guiding framework for unbiased physically-based rendering. We tackled the major challenges of neural-based online path guiding methods by proposing a novel closed-form density model, NASG. The simplicity and expressiveness of NASG allow it to be efficiently trained online via a tiny, MLP-based neural network with spare ray samples. 

 Through experiments, we show that under the same sample budget, our framework outperforms existing neural path guiding techniques and achieves comparable results to state-of-the-art statistical path guiding techniques. With proper implementation on GPU, our framework helps to improve the raw performance of a GPU path tracer via neural path guiding. This was achieved because our framework can effectively learn the spatial-varying distribution and guide a unidirectional path tracer with low overhead, allowing the path tracer to produce high-quality images with limited computational resources. Our work also shows that learning-based importance sampling has great potential in practical rendering tasks. We hope our work can help pave the path for research communities in industry and academia to adopt neural technologies for path tracing.
\section*{Acknowledgements}
This work was supported in part by JSPS KAKENHI Grant Number JP20K03551.
\appendix
\section{Continuity of NASG}
\label{sec:continuity}
The complete form of a NASG component is given by:
\begin{equation}
    \begin{aligned}
        &G(\boldsymbol{\mbv} ;[\boldsymbol{x,y,z}], \lambda, a, \epsilon, \kappa) \\
        &\quad= \begin{cases}
            \kappa \cdot \exp \left(2 \lambda\left(\frac{\boldsymbol{\mbv} \cdot \boldsymbol{z}+1}{2}\right)^{1+\epsilon+\frac{a(\boldsymbol{\mbv} \cdot \boldsymbol{x})^2}{1-\left(\boldsymbol{\mbv} \cdot \boldsymbol{z}\right)^2}}-2 \lambda\right)\left(\frac{\boldsymbol{\mbv} \cdot \boldsymbol{z}+1}{2}\right)^{\epsilon + \frac{a(\boldsymbol{\mbv} \cdot \boldsymbol{x})^2}{1-(\boldsymbol{\mbv} \cdot \boldsymbol{z})^2}} \\ \hspace{52mm} \text { if } \boldsymbol{\mbv} \neq \pm\boldsymbol{z} \\
            \kappa & \hspace{-20mm} \text { if } \boldsymbol{\mbv} = \boldsymbol{z} \\	
            0 & \hspace{-20mm} \text { if } \boldsymbol{\mbv} = -\boldsymbol{z}
         \end{cases}
    \end{aligned}
\end{equation}
where $\kappa$ is the lobe amplitude and $\epsilon$ is an auxiliary parameter introduced so that $G$ is continuous whenever $\epsilon > 0$. In practice, it is possible to set $\epsilon = 0$, although this breaks continuity at $\boldsymbol{\mbv} = -\boldsymbol{z}$ unless $a = 0$. For the rest of the content, we set $\kappa = 1$ and omit it from the notation.

\section{Derivation of the normalizing term}
\label{sec:integral}
Consider an NASG component defined as $G(\boldsymbol{\mbv} ;[\boldsymbol{x,y,z}], \lambda, a, \epsilon)$. Since $G$ is rotation invariant, we assume that $\bm{x} = (1, 0, 0)$, $\bm{y} = (0, 1, 0)$, and $\bm{z} = (0, 0, 1)$.
Then the surface integral of the function $G$ over $S^2$ 
is given by

\begin{equation}\label{eq:use spherical coordinates}
    \begin{aligned}
        & \int_{S^2} G(\bm{\mbv} ;[\boldsymbol{x,y,z}], \lambda, a, \epsilon)d\omega \\							
        & = \int_0^{2\pi}d\phi \int_0^{\pi} \exp\left( 2\lambda\left(\frac{\cos\theta + 1}{2} \right)^{1+\epsilon+a\cos^2\phi} -2\lambda \right) \\
        & \hspace{40mm} \left(\frac{\cos\theta + 1}{2}\right)^{\epsilon+a\cos^2\phi} \sin\theta d\theta.
    \end{aligned}
\end{equation}

We compute the inner integral over $\theta$ first.
Consider the following change of variable (for fixed $\phi$):
\begin{equation}
    \begin{aligned}
        &\frac{\cos\theta + 1}{2} = \left(\frac{\cos\tau + 1}{2}\right)^{\frac{1}{1+\epsilon+a\cos^2\phi}} \\
        \Leftrightarrow \ &\cos\theta = 2\left(\frac{\cos\tau + 1}{2}\right)^{\frac{1}{1+\epsilon+a\cos^2\phi}} -1,
    \end{aligned}
\end{equation}
where $0 \leqslant \tau \leqslant \pi$.
Then
\begin{equation}
    \begin{aligned}
        -\sin\theta d\theta = \frac{2}{1+\epsilon+a\cos^2\phi}
        \left(\frac{\cos\tau + 1}{2}\right)^{\frac{1}{1+\epsilon+a\cos^2\phi}-1}
        \left(-\frac{\sin\tau}{2}\right)d\tau,
    \end{aligned}
\end{equation}
so
\begin{equation}
    \begin{aligned}
        \sin\theta d\theta
        = \frac{1}{1+\epsilon+a\cos^2\phi}\left(\frac{\cos\tau + 1}{2}\right)^{-\frac{\epsilon+a\cos^2\phi}{1 +\epsilon + a\cos^2\phi}}\sin\tau d\tau,
    \end{aligned}
\end{equation}
and the inner integral with respect to $\theta$ becomes
\begin{equation}
    \begin{aligned}
        &\frac{1}{1+\epsilon+a\cos^2\phi}\int_0^\pi\exp(\lambda\cos\tau - \lambda)\sin\tau d\tau \\
        &= \frac{1}{1+\epsilon+a\cos^2\phi}\left[-\frac{\exp(\lambda\cos\tau - \lambda)}{\lambda}\right]_0^\pi \\
        &= \frac{1-e^{-2\lambda}}{\lambda(1+\epsilon+a\cos^2\phi)}.
    \end{aligned}
\end{equation}
Therefore, it follows that
\begin{equation}
    \begin{aligned}
        &\int_{S^2}G(\bm{\mbv} ;[\boldsymbol{x,y,z}], \lambda, a, \epsilon)d\omega \\
        &=\frac{1 - e^{-2\lambda}}{\lambda}\int_0^{2\pi}\frac{d\phi}{1+\epsilon+a\cos^2\phi} \\
        &=\frac{2(1-e^{-2\lambda})}{\lambda}\int_{-\frac{\pi}{2}}^{\frac{\pi}{2}}\frac{d\phi}{1+\epsilon+a\cos^2\phi}.
    \end{aligned}
\end{equation}
Now, consider the following change of variable:
\begin{equation}
    \begin{aligned}
        \tan\phi = \sqrt{\frac{1+\epsilon+a}{1+\epsilon}}\tan\rho.		
    \end{aligned}
\end{equation}
Then
\begin{equation}
    \begin{aligned}
        \frac{d\phi}{\cos^2\phi} = \sqrt{\frac{1+\epsilon+a}{1+\epsilon}}\frac{d\rho}{\cos^2\rho},	
    \end{aligned}
\end{equation}
where
\begin{equation}
    \begin{aligned}
        \cos^2\phi &= \frac{1}{1+\tan^2\phi}	\\
        &= \frac{1}{1+\frac{1+\epsilon+a}{1+\epsilon}\tan^2\rho} \\
        &= \frac{1+\epsilon}{\frac{1+\epsilon+a}{\cos^2\rho}-a} \\
        &= \frac{(1+\epsilon)\cos^2\rho}{1+\epsilon+a-a\cos^2\rho},
    \end{aligned}
\end{equation}
so
\begin{equation}
    \begin{aligned}
        d\phi = \frac{\sqrt{(1+\epsilon)(1+\epsilon+a)}}{1+\epsilon+a-a\cos^2\rho}d\rho.	
    \end{aligned}
\end{equation}
Consequently, it follows that
\begin{equation}\label{eq:K for general NASG}
    \begin{aligned}
        &\int_{S^2}G(\bm{v} ;[\boldsymbol{x,y,z}], \lambda, a, \epsilon)d\omega \\
        & =\frac{2(1-e^{-2\lambda})}{\lambda}\int_{-\frac{\pi}{2}}^{\frac{\pi}{2}}\frac{1}{1+\epsilon+\frac{(1+\epsilon)a\cos^2\rho}{1+\epsilon+a-a\cos^2\rho}}\cdot\frac{\sqrt{(1+\epsilon)(1+\epsilon+a)}}{1+\epsilon+a-a\cos^2\rho}d\rho \\
        & =\frac{2(1-e^{-2\lambda})}{\lambda}\int_{-\frac{\pi}{2}}^{\frac{\pi}{2}}\frac{d\rho}{\sqrt{(1+\epsilon)(1+\epsilon+a)}} \\
        & =\frac{2\pi(1-e^{-2\lambda})}{\lambda\sqrt{(1+\epsilon)(1+\epsilon+a)}}.
    \end{aligned}
\end{equation}
This expression reduces to \eqref{K for NASG} when $\epsilon=0$.

\section{Sampling NASG}
\label{sec:sampling}
Here, we discuss how to sample from the distribution on $S^2$ obtained by normalizing $G$ (cf.~\eqref{K for general NASG}). 
To this end, we essentially reverse the discussions in the previous section.
Define two functions
\begin{equation}
    \begin{aligned}
        &\Phi_E : [e^{-2\lambda}, 1] \times \left[-\frac{\pi}{2},\frac{\pi}{2}\right] 
        \rightarrow [0, \pi] \times \left[-\frac{\pi}{2},\frac{\pi}{2}\right], \\
        &\Phi_W : [e^{-2\lambda}, 1] \times \left[-\frac{\pi}{2}, \frac{\pi}{2}\right] 
        \rightarrow [0, \pi] \times \left[\frac{\pi}{2},\frac{3\pi}{2}\right]
    \end{aligned}
\end{equation}
by
\begin{equation}
    \begin{aligned}
        &\Phi_E(s, \rho) \\
        & =\left( \arccos\left(2\left(\frac{\log s}{2\lambda} + 1\right)^{\frac{1 + \epsilon+a - a\cos^2\rho}{(1+\epsilon)(1+\epsilon+a)}} - 1\right), \arctan\left(\sqrt{\frac{1+\epsilon+a}{1+\epsilon}}\tan\rho\right) \right) ,\\
        &\Phi_W(s, \rho) = \Phi_E(s, \rho) + (0, \pi),
    \end{aligned}
\end{equation}
where $(s, \rho) \in [e^{-2\lambda}, 1]\times[-\frac{\pi}{2}, \frac{\pi}{2}]$ (here, "E" and "W" stand for east and west, respectively). 

In practice, we just need to sample two uniform values $(\xi_0, \xi_1) \in [0,1]^2$ and linearly map them to $[e^{-2\lambda}, 1]\times[-\frac{\pi}{2}, \frac{\pi}{2}]$ to obtain $(s, \rho)$ used in the above equation.
We introduce another uniform random number $\xi_2 \in [0,1]$.
When $\xi_2 > 0.5$, we sample the eastern hemisphere and $\Phi_E(s,\rho)$ is the sampled direction's $(\theta, \phi)$. When $\xi_2\leqslant 0.5$, we instead use $\Phi_W(s,\rho)$.

Let us now argue that the above sampling method serves our purpose.
Observe that both $\Phi_E$ and $\Phi_W$ are bijections. Their inverses are given by
\begin{equation}
    \begin{aligned}
        &\Phi_E^{-1}(\theta, \phi) \\
        &= \left(\exp\left(2\lambda\left(\frac{\cos\theta+1}{2}\right)^{1+\epsilon+a\cos^2\phi}-2\lambda\right), \arctan\left(\sqrt{\frac{1+\epsilon}{1+\epsilon+a}}\tan\phi\right)\right) , \\
        &\Phi_W^{-1}(\theta, \phi) = \Phi_E^{-1}(\theta, \phi - \pi),
    \end{aligned}
\end{equation}
where $(\theta, \phi) \in [0, \pi]\times[-\frac{\pi}{2}, \frac{\pi}{2}]$ for the former, and $(\theta, \phi) \in [0, \pi]\times[\frac{\pi}{2}, \frac{3\pi}{2}]$ for the latter.
The Jacobian for $\Phi_E^{-1}$ is computed as
\begin{equation}
    \begin{aligned}
        &J_{\Phi_E^{-1}}(\theta, \phi)  \\
        &= \exp\left(2\lambda\left(\frac{\cos\theta + 1}{2}\right)^{1+\epsilon+a\cos^2\phi} - 2\lambda\right)	\\
        &\times 2\lambda(1 + \epsilon+a\cos^2\phi)\left(\frac{\cos\theta+1}{2}\right)^{\epsilon+a\cos^2\phi}\left(-\frac{\sin\theta}{2}\right) \\
        &\times \frac{1}{1 + \frac{1+\epsilon}{1+\epsilon+a}\tan^2\phi}\sqrt{\frac{1+\epsilon}{1+\epsilon+a}}\frac{1}{\cos^2\phi} \\
        &= \lambda \sqrt{(1+\epsilon)(1+\epsilon+a)}\exp\left(2\lambda\left(\frac{\cos\theta+1}{2}\right)^{1+\epsilon+a\cos^2\phi} - 2\lambda\right) \\
        & \times \left(\frac{\cos\theta + 1}{2}\right)^{\epsilon+a\cos^2\phi}(-\sin\theta).
    \end{aligned}
\end{equation}

Let $X$ be a random variable on $S^2$, which we also view as a function of $(\theta,\phi)$.
Then, the expected value of $X$ over the points $\Phi_E(s,\rho)$, where the $(s,\rho)$ are uniformly sampled from $[e^{-2\lambda}, 1] \times [-\frac{\pi}{2}, \frac{\pi}{2}]$, is given by

\begin{equation}
    \begin{aligned}
        \mathbb{E}\left[X\circ\Phi_E\right] 
        = \frac{1}{\pi(1-e^{-2\lambda})}\int_{[e^{-2\lambda}, 1]\times[-\frac{\pi}{2}, \frac{\pi}{2}]} X \circ \Phi_E(s, \rho)dsd\rho.
    \end{aligned}
\end{equation}
By taking the change of variables $(s, \rho) = \Phi_E^{-1}(\theta, \phi)$, we have
\begin{equation}
    \begin{aligned}
        &\mathbb{E}\left[X\circ\Phi_E\right] \\
        &= \frac{1}{\pi(1 - e^{-2\lambda})}\int_{[0, \pi]\times[-\frac{\pi}{2}, \frac{\pi}{2}]}X(\theta, \phi)\left|J_{\Phi_E^{-1}}(\theta, \phi)\right|d\theta d\phi \\
        &= \frac{\lambda\sqrt{(1+\epsilon)(1+\epsilon+a)}}{\pi(1-e^{-2\lambda})} \\
        & \int_{[0, \pi]\times[-\frac{\pi}{2}, \frac{\pi}{2}]}X(\theta, \phi) 
         \exp\left(2\lambda\left(\frac{\cos\theta+1}{2}\right)^{1+\epsilon+a\cos^2\phi} - 2\lambda\right) \\
        & \hspace{37mm} \left(\frac{\cos\theta+1}{2}\right)^{\epsilon+a\cos^2\phi}\sin\theta d\theta d\phi.
    \end{aligned}
\end{equation}
We also obtain a similar expression for $\mathbb{E}\left[X\circ\Phi_W\right]$.
Since we choose each of the eastern and western hemispheres with probability $\frac{1}{2}$ according to the values of $\xi_2$, the overall expected value becomes
\begin{equation}
    \begin{aligned}
        \frac{1}{2}&\mathbb{E}\left[X\circ\Phi_E\right] +\frac{1}{2} \mathbb{E}\left[X\circ\Phi_W\right] \\
        &= \frac{\lambda\sqrt{(1+\epsilon)(1+\epsilon+a)}}{2\pi(1-e^{-2\lambda})}
        \int_{S^2}X(\bm{v})G(\bm{v} ;[\boldsymbol{x,y,z}], \lambda, a, \epsilon)d\omega.
    \end{aligned}
\end{equation}
See \eqref{use spherical coordinates}.
It follows that our sampled directions obey the distribution obtained by normalizing $G$, as desired.
	
\bibliographystyle{ACM-Reference-Format}
\bibliography{OLLF}
\end{document}